\documentclass[sort&compress,times]{elsarticle}
\usepackage[T1]{fontenc}
\usepackage[utf8]{inputenc}
\usepackage{url}
\usepackage[breaklinks=true]{hyperref}\hypersetup{pdfborder={0 0 0}}
\usepackage{amsmath,amssymb,mathtools}
\usepackage{color}
\usepackage{graphicx}
\usepackage{caption}
\usepackage{subcaption}
\usepackage[algo2e,noend,linesnumbered]{algorithm2e}
\makeatletter\renewcommand{\@algocf@capt@plain}{above}\makeatother
\usepackage{tikz}

\newcommand{\TD}{\ensuremath{\mathit{TD}}}
\newcommand{\SSQ}{\ensuremath{\mathit{SSQ}}}
\newcommand{\dist}{\ensuremath{\mathit{d}}}
\DeclareMathOperator*{\argmax}{arg\,max}
\DeclareMathOperator*{\argmin}{arg\,min}
\newcommand{\kmedoids}{\mbox{$k$-medoids}}
\newcommand{\change}{\ensuremath{\Delta}}%
\newcommand{\best}[1]{{#1\mathstrut}^*\!}
\newcommand{\Change}{\Delta\TD}
\newcommand{\nearest}{\text{nearest}}

\newcommand{\reffig}[1]{Figure~\ref{#1}}
\newcommand{\refsec}[1]{Section~\ref{#1}}
\newcommand{\refalg}[1]{Algorithm~\ref{#1}}
\newcommand{\refeqn}[1]{Equation~\ref{#1}}
\newcommand{\reftab}[1]{Table~\ref{#1}}

\title{Fast and Eager $k$-Medoids Clustering:\\
$O(k)$ Runtime Improvement of the\\ PAM, CLARA, and CLARANS Algorithms\tnoteref{tref1}}
\author[1]{Erich Schubert\corref{cor1}\fnref{fn1}}
\address[1]{TU Dortmund University, Informatik VIII, 44221 Dortmund, Germany}
\ead[1]{erich.schubert@tu-dortmund.de}

\author[2]{Peter J.{} Rousseeuw}
\address[2]{Statistics and Data Science Section, KU Leuven, Celestijnenlaan 200b - box 2400, 3001 Leuven, Belgium}
\ead[2]{peter@rousseeuw.net}

\tnotetext[tref1]{This is an extended version of \cite{DBLP:conf/sisap/SchubertR19} presented at the SISAP'19 conference.}
\fntext[fn1]{
Part of the work on this paper has been supported by Deutsche Forschungsgemeinschaft (DFG)
-- project number 124020371 --
within the Collaborative Research Center SFB 876 ``Providing Information by Resource-Constrained Analysis'', project A2.
\url{https://sfb876.tu-dortmund.de/}}
\cortext[cor1]{Corresponding author}

\begin{document}

\begin{abstract}
Clustering non-Euclidean data is difficult, and one of the most used
algorithms besides hierarchical clustering is the popular algorithm
Partitioning Around Medoids (PAM), also simply referred to as $k$-medoids clustering.
In Euclidean geometry the mean---as used in $k$-means---is a good estimator for the cluster
center, but this does not exist for arbitrary dissimilarities. PAM uses the
medoid instead, the object with the smallest dissimilarity to all others
in the cluster. This notion of centrality can be
used with any (dis-)similarity, and thus is of high relevance to many
domains and applications.
A~key issue with PAM is its high run time cost.
We propose modifications to the PAM algorithm
that achieve an $O(k)$-fold speedup in the second (``SWAP'') phase
of the algorithm, but will still find the same results as the original PAM algorithm.
If we relax the choice of swaps performed (while retaining comparable
quality), we can further accelerate the algorithm by eagerly performing
additional swaps in each iteration.
With the substantially faster SWAP, we can now explore
faster initialization strategies, because (i)~the classic (``BUILD'')
initialization now becomes the bottleneck, and (ii)~our swap is fast enough
to compensate for worse starting conditions.
We also show how the CLARA and CLARANS
algorithms benefit from the proposed modifications.
While we do not study the parallelization of our approach in this work, it
can easily
be combined with earlier approaches to use PAM and CLARA on big data (some of
which use PAM as a subroutine, hence can immediately benefit from these improvements),
where the performance with high $k$ becomes increasingly important.
In experiments on real data with $k = 100,200$, we observed a $458\times$ respectively
$1191\times$ speedup compared to the original PAM SWAP algorithm,
making PAM applicable to larger data sets, %
and in particular to higher~$k$.

\end{abstract}

\maketitle

\begin{tikzpicture}[overlay, remember picture]
\node[red, font=\small, yshift=-2cm, anchor=north, text width=12cm] at (current page.north) {
Published as:

Erich Schubert, Peter J. Rousseeuw,
Fast and eager k-medoids clustering: \\
O(k) runtime improvement of the PAM, CLARA, and CLARANS algorithms,\\
Information Systems, 2021,
\url{https://doi.org/10.1016/j.is.2021.101804}.};
\end{tikzpicture}

\section{Introduction}

Clustering is a common unsupervised machine learning task, in which the data set
has to be automatically partitioned into ``clusters'', such that objects within
the same cluster are more similar, while objects in different clusters are
more different.
There is not (and likely never will be) a generally accepted definition
of a cluster~\citep{DBLP:journals/ibmrd/Bonner64},
because ``clusters are, in large part, in the eye of the
beholder''~\citep{DBLP:journals/sigkdd/Estivill-Castro02}, meaning that
every user may have different enough needs and intentions to want a
different algorithm and notion of cluster.
And therefore, over many years of research,
hundreds of clustering algorithms
and evaluation measures have been proposed, each with their merits and drawbacks.
Nevertheless, a few seminal methods such as hierarchical clustering, $k$-means,
PAM~(Partitioning Around Medoids, \citealt{KauRou87,KauRou90a}), and
\mbox{DBSCAN}~(Density-Based Spatial Clustering of Applications with Noise, \citealt{DBLP:conf/kdd/EsterKSX96})
have received repeated and widespread use. One may be tempted
to think that %
these \emph{classic} methods have all been well researched and
understood, but there are still many scientific publications 
trying to explain these algorithms better (e.g.,~\citealp{DBLP:journals/tods/SchubertSEKX17}),
trying to parallelize and scale them to larger data sets (e.g.,~\citealp{DBLP:journals/datamine/LijffijtPP15,serss/YangL14}),
trying to better understand similarities and relationships among the published methods
(e.g.,~\citealp{DBLP:conf/lwa/SchubertHM18}),
or proposing further improvements --
and so does this paper for the widely used PAM algorithm, also often referred to as \kmedoids{} clustering.

In hierarchical agglomerative clustering (HAC), each object is initially its
own cluster. The two closest clusters are then merged repeatedly to build a cluster
tree called dendrogram.
HAC is a very flexible method: it can be used with any
distance or (dis-)similarity, and it allows for different rules of aggregating
the object distances into cluster distances, such as the minimum (``single linkage''),
average, or maximum (``complete linkage'').
Single linkage directly corresponds to the minimum spanning tree of the distance graph.
While the dendrogram is a powerful visualization for small data sets, extracting flat
partitions from hierarchical clustering is not trivial, and thus users often turn to simpler methods.

A classic method taught in textbooks
is $k$-means (for an overview of
the complicated history of $k$-means, refer to~\citealp{Bock07}), where the data is
modeled using $k$ cluster means, that are iteratively refined by assigning all objects
to the nearest mean, then recomputing the mean of each cluster.
This converges to a local optimum because the mean is the least squares estimator of location,
and both steps reduce the same quantity, a measure known as sum-of-squared errors~(\SSQ{}): %
\begin{align}
\SSQ :=&
\textstyle
\sum\nolimits_{i=1}^k \sum\nolimits_{x_c \in C_i} ||x_c - \mu_i||_2^2
\,.
\label{eqn:ssq}
\end{align}

In \kmedoids{}, the data is modeled similarly, using $k$ representative objects $m_i$
called medoids (chosen from the data set; defined below) that serve as
prototypes for the clusters instead of means
in order to allow using arbitrary other dissimilarities and
arbitrary input domains (not restricted to vector spaces),
using the absolute error criterion (``total deviation'', \TD) as objective:
\begin{align}
\TD :=&
\textstyle
\sum\nolimits_{i=1}^k \sum\nolimits_{x_c \in C_i} \dist(x_c, m_i)
\,,
\label{eqn:td}
\end{align}
\noindent
which is the sum of dissimilarities of each point $x_c\in C_i$ to the medoid~$m_i$ of its cluster.
If we use squared Euclidean as distance function (i.e., $d(x,m)=||x-m||_2^2$),
we almost obtain the usual \SSQ{} objective used by $k$-means, except that $k$-means is free to choose any
$\mu_i\in\mathbb{R}^d$, whereas in \kmedoids{} $m_i\in C_i$ must be one of the original
data points.
But on the other hand, the \kmedoids{} objective can be used with any distance function,
even when our data is is not a $\mathbb{R}^d$ vector space.
For \emph{squared} Euclidean distances and Bregman divergences,
the arithmetic mean is the optimal choice for $\mu$. %
For $L_1$ distance (i.e, $\sum |x_i-y_i|$), also called Manhattan distance,
the component-wise median is a better choice in $\mathbb{R}^d$~\citep%
{DBLP:conf/nips/BradleyMS96}. For unsquared Euclidean distances,%
\footnote{It is a common misconception that $k$-means would minimize Euclidean distances:
It optimizes the sum of \emph{squared} Euclidean distances, which is not equivalent, and even then the textbook
algorithm may end up slightly off a local optimum, because always assigning a point to its nearest center \emph{can}
increase the variance by moving the centers away from other points \citep{HartiganW79}.}
we get the much harder Weber problem~\citep{DBLP:journals/mp/Overton83},
which has no closed-form exact solution~\citep{DBLP:conf/nips/BradleyMS96}.
For a recent survey of algorithms for the Weber point see \cite{FritzFC12}.
For other distance functions, finding a closed form to compute the best $m_i$ would require
a separate non-trivial mathematical analysis of each distance function separately.
Furthermore, our input domain is not necessarily a $\mathbb{R}^d$ vector space.
In \kmedoids{} clustering, we therefore constrain $m_i$ to be one of our data samples.
The medoid of a set $C$ is defined as the object with the smallest sum of dissimilarities
(or, equivalently, smallest average)
to all other objects in the set:
\begin{align}
\operatorname{medoid}(C) := \argmin\nolimits_{x_m\in C} \textstyle\sum\nolimits_{x_c\in C} \dist(x_c, x_m)
\,.
\label{eqn:def-medoid}
\end{align}
This definition does not require the dissimilarity to be a metric,
and by using $\argmax$ %
it can also be applied to similarities.
The algorithms discussed in detail in this article all can trivially be modified to maximize similarities
rather than minimizing distances, and none assumes the triangular inequality.
Partitioning Around Medoids (PAM, \citeauthor{KauRou87}, \citeyear{KauRou87,KauRou90a})
is the most widely known clustering algorithm to
find a good partitioning using medoids, with respect to $\TD$ (\refeqn{eqn:td}).

\medskip
\begingroup\leftskip2em\rightskip\leftskip\noindent
This is an extension of earlier work presented at the SISAP'19 conference:
\smallskip\par\noindent
Schubert, Erich, Rousseeuw, Peter J., 2019. Faster $k$-medoids clustering:
Improving the PAM, CLARA, and CLARANS algorithms.
\\
In:
Similarity Search and Applications. SISAP~2019.
pp.~171-187.
\\
doi: \href{https://doi.org/10.1007/978-3-030-32047-8_16}{10.1007/978-3-030-32047-8\_16}.
\par\endgroup
\medskip

\noindent
In comparison to the original conference version, the improved version presented here
modifies the algorithm in a way that allows to prove the speedup factor of $O(k)$
compared to the original PAM algorithm by completely eliminating the nested loop of length~$k$.
We furthermore study a new variant using eager swapping that further improves runtime.
We also include a brief recap on the history of the PAM algorithm,
updated and more extensive benchmarks on additional data sets,
and cover additional related work.

\subsection{On the History of PAM}

In the early eighties many clustering methods were restricted to dealing with metric data,
i.e., coordinates of geometric points.
The PAM algorithm was developed at that time (but only published later).
PAM was part of a project to construct clustering methods that could deal
with arbitrary dissimilarity matrices (subjective judgments, confusion matrices, \ldots)
that did not even have to satisfy the triangle inequality.
\citet{KauRou87} proposed the name \emph{medoid} for an object with lowest
total dissimilarity to the other objects of its cluster,
in order to distinguish it from the \emph{(geometric) median} for metric data
(for a survey see \citealp{FritzFC12}),
defined by minimizing the sum of Euclidean distances over all geometric points,
not only the data points.
Of course PAM can handle metric data as well by first computing a dissimilarity
matrix from them, e.g., using Euclidean or Manhattan distance.
The program DAISY (an anagram of DISsimilAritY, \citealt{KauRou90})
also computed dissimilarity measures for data with non-numerical variables.

Originally PAM was run on the first generation of IBM PC's that only had two floppy disks of 360KB each
(the left one containing the DOS operating system), 64kB of internal memory, and no hard drive.
The Fortran code of PAM could only be compiled after splitting it in pieces.
At that time the main limitation of PAM was not so much its computation time but mainly
the $O(n^2)$ memory required, allowing to analyze datasets with up to about $n=150$ cases only.
This restriction was then circumvented by the program CLARA~(Clustering LARge Applications, \citealt{KaufmanR86,KauRou90b}).

As for the naming of the k-medoid algorithm, it was first thought to combine the initials
of K-Medoid into KIM, after the daughter of one of the authors.
But this only matched two out of three letters.
Thinking a bit longer led to the descriptive name Partitioning Around Medoids with abbreviation PAM,
then the well-known name of a character played by Victoria Principal in the television series \emph{Dallas}.
In fact all algorithms in the subsequent book \citep{KauRou90} were given female abbreviations,
with for instance DAISY inspired by HAL's song in Kubrick's film \emph{2001:~A~Space Odyssey}.

Around the same time, a family of related problems received a lot of attention
in a different domain, trying to find the optimal matching of consumer locations
and potential facility locations.
But of course finding related work was, at that time, much harder than it is today
with electronic access, Wikipedia, and full text search engines.
Today, we can more easily find such connections across different domains.

\subsection{Related Location-Allocation Problems}

Closely related approaches and problems can be found in other domains such as in
operations research and management science
with different names such as ``$p$-medians'' (not to be confused
with geometric medians). The $k$-medoids clustering problem can be seen as a
\emph{symmetric} and \emph{discrete} special case of $p$-medians, and
the uncapacitated facility location problem (UFLP),
where the number of facilities to be opened, $k$~respectively~$p$, is constant;
where all facilities have the same opening costs and
where all customers have equal demand.
A survey and annotated bibliography of the $p$-median problem and some of its variations
in facility location can be found in \citet{DBLP:journals/networks/Reese06}.

In such domains, authors such as \cite{DBLP:journals/ior/TeitzB68}
and \cite{DBLP:journals/ibmsj/Maranzana63} have considered various heuristics
for this version of the problem.
Parts of the PAM algorithm can be found in this literature by the name ``greedy''
for the BUILD initialization of PAM; ``interchange'' or ``vertex substitution''
for the SWAP part of PAM; and ``alternate'' for the $k$-means-style iteration
technique also discussed in data mining literature (e.g., \citealt{DBLP:journals/eswa/ParkJ09,DBLP:books/sp/HastieFT01}).
Several variations have been suggested, such as the ``fast interchange''
heuristic of \citet{journals/infor/Whitaker83}.
\citet{journals/ejor/Beasley85} used a Cray-1S supercomputer to
find the exact solution for such problems with up to 900 instances with
a branch-and-bound approach. For some of the ORlib problems, the exact solutions
could still not be determined within 600 seconds at that time.
Because of these similarities, our algorithms may be of interest for researchers
from these domains too, as it should be possible to add support for varying demand
and asymmetric problems (possibly even for capacitated facility location),
and retain the $O(k)$ speedup over the standard local search heuristic popular in these
domains, at least for the initialization of more complex search methods.

In \refsec{sec:pam} we first introduce the original PAM algorithm and important variants.
We then discuss the bottleneck of the algorithm, finding the best swap, in \refsec{sec:best-swap} and
introduce our improvements to the algorithm.
In \refsec{sec:experiments} we perform an experimental performance evaluation on standard benchmark
data sets and investigate empirical scalability in different parameters.
We provide an outlook for future research in \refsec{sec:outlook} and conclusions in \refsec{sec:conclusions}.

\section{Partitioning Around Medoids (PAM) and its Variants}\label{sec:pam}

The ``Program PAM''~\citep{KauRou87,KauRou90a} consists of two algorithms,
BUILD to choose an initial clustering, and SWAP
to improve the clustering towards a local optimum (finding
the global optimum of the $k$-medoids problem is, unfortunately, NP-hard as shown by
\citealp{journals/jam/KarivH79}).
The algorithms require a dissimilarity matrix
(for example computed using the routine DAISY of \citealp{KauRou90}),
which requires $O(n^2)$ memory and
typically for many popular distance functions in $d$ dimensional data
$O(n^2 d)$ time to compute (but potentially
much more for expensive distances such as earth mover's distance also known as Wasserstein metric).
Computing the distance matrix often is already a bottleneck in many cases.

\begin{algorithm2e}[b!]\addtolength{\hsize}{1.5em}
\caption{PAM BUILD: Find initial cluster centers.}
\label{alg:build}
\SetKw{Return}{return}
$(\TD,m_1) \leftarrow (\infty, \text{null})$\;
\ForEach(\tcp*[f]{First medoid}){$x_c$}{
  $\TD_j\leftarrow 0$\;
  \lForEach{$x_o \neq x_c$}{
    $\TD_j \leftarrow \TD_j + \dist(x_o,x_c)$%
  }
  \lIf(\tcp*[f]{Smallest distance sum}){$\TD_j<\TD$}{
    $(\TD, m_1) \leftarrow (\TD_j, x_c)$%
  }
}
\ForEach(\tcp*[f]{Initialize distance\ \ \scriptsize$\mathllap{\hookleftarrow}$}\label{line:build-init}){$x_o \neq m_1$}{
  $\dist_{\textit{nearest}}(o) \leftarrow \dist(m_1,x_o)$%
  \tcp*[r]{to nearest medoid\ \ \ \ }
}
\For(\tcp*[f]{Other medoids}){$i=1\ldots k-1$}{
  $(\best{\Change}, \best{x}) \leftarrow (\infty, \text{null})$\;
  \ForEach{$x_c \not\in \{m_1,\ldots,m_i\}$}{
    $\Change\leftarrow 0$\;
    \ForEach{$x_o \not\in \{m_1,\ldots,m_i,x_c\}$}{
      $\delta \leftarrow \dist(x_o,x_c) - \dist_{\textit{nearest}}(o)$%
      \tcp*[r]{Reduction in \TD}\label{line:build-cached}
      \lIf{$\delta<0$}{$\Change\leftarrow\Change + \delta$}
    }
    \lIf(\tcp*[f]{Best reduction}){$\Change<\best{\Change}$}{
      $(\best{\Change}, \best{x}) \leftarrow (\Change, x_c)$%
    }
  }
  $(\TD, m_{i+1}) \leftarrow (\TD + \best{\Change}, \best{x})$\;
  \ForEach(\tcp*[f]{Update distances\ \ \scriptsize$\mathllap{\hookleftarrow}$}){$x_o \not\in \{m_1,\ldots ,m_{i+1}\}$}{
    $\dist_{\textit{nearest}}(o) \leftarrow \min\{\dist_{\textit{nearest}}(o), \dist(x_o,m_{i+1})\}$%
    \tcp*[r]{to nearest medoid\ }\label{line:build-update}
  }
}
\Return $\TD, \{m_1,\ldots,m_k\}$\;
\end{algorithm2e}

In order to find a good initial clustering
(rather than relying on a random sampling strategy as commonly used with $k$-means),
BUILD chooses $k$~times the point which yields the smallest distance sum \TD{}
(this means first choosing the point with the smallest distance
to all others; afterwards always adding the point that reduces \TD{} most).
We give a pseudocode in \refalg{alg:build}, where we use $\Change$ as symbol for the
change in $\TD$ (which should be negative to be beneficial),
and add an asterisk $\best{}$ for the best values found so far.
A subtle but important detail used in BUILD
(independently suggested by \citealt{journals/infor/Whitaker83} in the ``fast greedy'' approach)
is to cache the distance to the nearest medoid in line~\ref{line:build-init},
then use this in the loop in line~\ref{line:build-cached}, and update it when a new
medoid has been chosen in line~\ref{line:build-update}. This avoids an additional nested loop over $k$
inside the computation of $\Change$,
and reduces the runtime of the naive implementation from $O(n^2 k^2)$
to $O(n^2 k)$ time. Nevertheless, this remains a fairly expensive algorithm.
The motivation here was to find a good starting point, in order to require fewer
iterations of the refinement procedure.
In the experiments, we will also study whether
randomized initialization approaches
are an interesting alternative.
\begin{algorithm2e}[b!]\addtolength{\hsize}{1.5em}
\caption{PAM SWAP: Iterative improvement.}
\label{alg:swap}
\SetKwFor{Repeat}{repeat}{}{end}
\SetKw{StopIf}{break loop if}
\lForEach{$x_o$}{
  compute $\nearest(o), \dist_{\nearest}(o), \dist_{\text{second}}(o)$\label{line:swap-nearest}%
}
\Repeat(\label{line:swap-outer-loop}){}{
  $(\best{\Change}, \best{m}, \best{x}) \leftarrow (0, \text{null}, \text{null})$\;
  \ForEach(\tcp*[f]{each medoid\ \ \ \ }){$m_i \in \{m_1,\ldots,m_k\}$}{
    \ForEach(\tcp*[f]{each non-medoid}){$x_c \not\in \{m_1,\ldots,m_k\}$}{
      $\Change\leftarrow 0$\;
      \ForEach{$x_o\not\in \{m_1,\ldots,m_k\}\setminus m_i$}{
        $\Change \leftarrow\Change + \change(x_o,m_i,x_c)$%
        \tcp*[r]{compute loss change}%
        \label{line:swap-change}
      }
      \If(\tcp*[f]{new best swap found}){$\Change < \best{\Change}$}{
        $(\best{\Change}, \best{m}, \best{x}) \leftarrow (\Change, m_i, x_c)$\;
      }
    }
  }
  \StopIf $\best{\Change}\geq 0$\;
  swap roles of medoid $\best{m}$ and non-medoid $\best{x}$\tcp*{perform best swap}\label{line:swap-swap}
  \lForEach{$x_o$}{
    update $\nearest(o), \dist_{\nearest}(o), \dist_{\text{second}}(o)$\label{line:swap-update}%
  }
  $\TD \leftarrow \TD + \best{\Change}$\;
}
\Return $\TD,\{m_1,\ldots,m_k\}$\;
\end{algorithm2e}

The second part of PAM, which is the main focus of this paper, was named SWAP.
It improves the clustering by considering all possible simple changes to the set of $k$~medoids,
which effectively means replacing (swapping) some medoid with some non-medoid, which gives
$k\cdot(n-k)$ candidate swaps.
If it reduces \TD{}, the best such change is then applied, in the spirit of a steepest-descent method,
and this process is repeated until no further improvements are found. The process then has
reached a local (but not necessarily the global) optimum where no solution that agrees on $k-1$ medoids is better.
Because the possible search space of medoids is finite (although large: $\smash{\binom{n}{k}}$),
a steepest-descent method must converge with a finite number of iterations.

We give a pseudocode of this in \refalg{alg:swap}.
In line \ref{line:swap-nearest} we compute---and cache---the
necessary data to compute the $\change(x_o,m_i,x_c)$ function
(\refeqn{eqn:change}, explained in \refsec{sec:best-swap})
in line~\ref{line:swap-change} efficiently.
These values then need to be updated after performing a swap in line~\ref{line:swap-update}.
With the cached values, the run time of the main loop of this algorithm is
$O(k (n-k)^2)$ for each iteration. %
(A similar optimization can also be found in \citealp*{journals/infor/Whitaker83}.)
While the authors of PAM assumed that only few iterations will be needed (if the algorithm
is already initialized well, using the BUILD algorithm above),
we do see an increasing number of iterations
with increasing amounts of data and increasing $k$ and cannot give a non-trivial bound for the maximum
number of iterations necessary,
but usually we observe fewer than $k$ iterations (as also observed by \citealp*{journals/infor/Whitaker83}).

Both the pseudocode for BUILD and SWAP given here omit the details of managing the cached distances.
For each object, we need to store the index of the nearest medoid $\textit{nearest}(o)$
and its distance $\dist_{\textit{nearest}}(o)$, and also the distance to the second nearest medoid
$\dist_{\textit{second}}(o)$ (for brevity, we will also use $\dist_n(o)$ and $\dist_s(o)$
in some places, because of space restrictions).
In line~\ref{line:swap-update} (or better during line~\ref{line:swap-swap}, when executing the best swap),
we need to carefully update these cached values
(if we additionally store the index of the second nearest center,
we may be able to avoid some more distance computations if the new medoid becomes the
nearest or second nearest after a swap).

\subsection{Variants and Extensions of PAM}

The algorithm CLARA~(Clustering LARge Applications, \citealt{KaufmanR86,KauRou90b}) repeatedly applies PAM on a subsample with
$n^\prime\!\ll\!n$ objects, with the suggested value $n^\prime\!=\!40+2k$.
Afterwards, the remaining objects are assigned to their closest
medoid. The run with the least $\TD$ (on the entire data) is returned. If the sample size
is chosen $n^\prime\in O(k)$ as suggested, the run time reduces to about
$O(k^3+n)$, which explains why the approach
is typically used only with small $k$~\citep{LucasiusDK93}.
Because CLARA uses PAM internally,
it will directly benefit from our improvements proposed in this article.

\cite{LucasiusDK93} propose a genetic algorithm for \kmedoids{},
by performing a randomized exploration of the search space
based on ``mutation'' of the best solutions found so far.
Crossover mutations correspond to taking some medoids from both ``parents'',
whereas mutations replace medoids with random objects.
It is not obvious that this will efficiently provide a sufficient coverage of the enormous
search space (there are $\tbinom{n}{k}\!=\!\frac{n!}{k!(n-k)!}$ possible sets of medoids) for a large $k$.
In order to benefit from the proposed improvements, a more systematic mutation strategy would
need to be adopted, making the method similar to CLARANS below.

\cite{DBLP:journals/eswa/WeiLH03} found the genetic methods to work
only for small data sets, small $k$, and well separated symmetric clusters,
and it was usually outperformed by CLARANS.
The algorithm CLARANS~(Clustering Large Applications based on RANdomized Search, \citealt{DBLP:conf/vldb/NgH94,DBLP:journals/tkde/NgH02})
interprets the search space as a high-dimensional hypergraph,
where each edge corresponds to swapping a medoid and non-medoid. On this graph it performs a
randomized greedy exploration, where the first edge that reduces the loss $\TD$ is followed
until no edge can be found with $p\!=\!1.25\% \cdot k(n-k)$ attempts.
In \refsec{sec:better-clarans} we will outline how our approach can be used to explore the
$k$~edges corresponding to all medoids at a time efficiently;
this will allow exploring a larger part of the search space in similar time,
but we expect the savings to be relatively small compared to the improvements gained in PAM.

Other proposals include optimizations for Euclidean space
(e.g.,~\citealp{DBLP:journals/datamine/Estivill-CastroY04,DBLP:journals/algorithmica/Estivill-CastroH01}),
simulated annealing \citep{DBLP:journals/heuristics/MurrayC96},
variable neighborhood search \citep{DBLP:journals/cor/MladenovicH97},
and tabu search heuristics
(e.g.,~\citealp{journals/ejor/RollandSC97}).
\citet{DBLP:conf/pakdd/Estivill-CastroM98} suggested stopping early
when observing diminishing returns with a ``fast interchange'' heuristic as used by
\citet{journals/infor/Whitaker83}.
\citet{DBLP:conf/aistats/NewlingF17} propose an interesting sub-quadratic algorithm,
but it requires the distances to be metric (an additional problem is discussed
in the following section).
For a broad survey of related techniques in operations research,
see \citet{DBLP:journals/networks/Reese06}.

\cite{DBLP:journals/jmma/ReynoldsRIR06} discuss an interesting trick to speed
up PAM. They show how to decompose the change in the loss function into two components,
where the first depends only on the medoid removed, the second part only on the new point.
This decomposition forms the base for our approach, and we will thus discuss it in
\refsec{sec:best-swap} in more detail. 

\subsection{Alternating \kmedoids{} Algorithm}

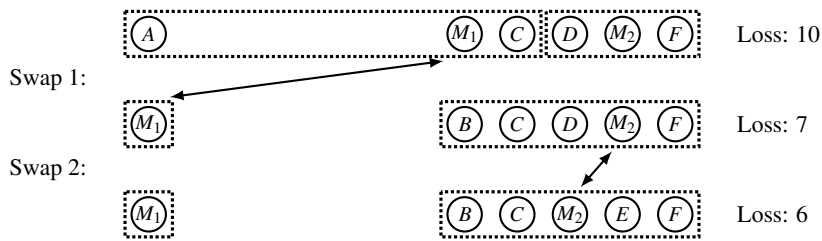
\begin{figure}\centering
\begin{tikzpicture}[xscale=.7,yscale=.6,font=\small]
\tikzstyle{n}=[draw,circle,thick,inner sep=-5pt, minimum width=1.3em,font=\footnotesize]
\tikzstyle{b}=[very thick, densely dotted]
\path[use as bounding box] (-8,.4) rectangle (+6,-4.4);

\draw[b] (-6.45,.5) rectangle (1.45,-.5);
\draw[b] (1.55,.5) rectangle (4.45,-.5);
\node[n] at (-6,0) {$A$};
\node[n] at (0,0) {$M_1$};
\node[n] at (1,0) {$C$};
\node[n] at (2,0) {$D$};
\node[n] at (3,0) {$M_2$};
\node[n] at (4,0) {$F$};

\node[anchor=west] at (5,0) {Loss: 10};
\node[anchor=east] at (-7,-1) {Swap 1:};

\draw[thick,<->,>=latex] (-.4,-.6) -- (-5.6,-1.4);

\draw[b] (-6.45,-1.5) rectangle (-5.55,-2.5);
\draw[b] (-.45,-1.5) rectangle (4.45,-2.5);
\node[n] at (-6,-2) {$M_1$};
\node[n] at (0,-2) {$B$};
\node[n] at (1,-2) {$C$};
\node[n] at (2,-2) {$D$};
\node[n] at (3,-2) {$M_2$};
\node[n] at (4,-2) {$F$};

\node[anchor=west] at (5,-2) {Loss: 7};
\node[anchor=east] at (-7,-3) {Swap 2:};

\draw[thick,<->,>=latex] (2.8,-2.6) -- (2.2,-3.4);

\draw[b] (-6.45,-3.5) rectangle (-5.55,-4.5);
\draw[b] (-.45,-3.5) rectangle (4.45,-4.5);
\node[n] at (-6,-4) {$M_1$};
\node[n] at (0,-4) {$B$};
\node[n] at (1,-4) {$C$};
\node[n] at (2,-4) {$M_2$};
\node[n] at (3,-4) {$E$};
\node[n] at (4,-4) {$F$};

\node[anchor=west] at (5,-4) {Loss: 6};

\end{tikzpicture}
\caption{Problematic example for the alternating heuristic: %
the $k$-means style approach is stuck in the top solution, %
while the SWAP heuristic can reach better solutions by reassigning points during the swap.}
\label{fig:alternating}
\end{figure}

\cite{DBLP:journals/eswa/ParkJ09} propose a ``$k$-means like'' algorithm for \kmedoids{}
that is supposedly ``simple and fast''
(actually this was already considered before by, e.g.,
\citealp{DBLP:journals/ibmsj/Maranzana63}; \citealp{DBLP:books/sp/HastieFT01}; \citealp{DBLP:journals/jmma/ReynoldsRIR06}).
\citet{DBLP:conf/aistats/NewlingF17} propose a sub-quadratic variant for this algorithm
for metric data, but unfortunately do not compare the result quality to alternatives
(in \citealp{DBLP:conf/nips/NewlingF17}, the authors observed that CLARANS produced
much better results than this approach, in neither work they included PAM).
This approach is well known in operations research literature by the name ``alternate''
because it consists of two alternating steps,
where in each iteration the medoid is chosen to be the object with the smallest distance sum to
other members of the cluster, then each point is assigned to the nearest medoid until
\TD{} no longer decreases.
Choosing cluster medoids with a $k$-means like strategy takes $O(n^2)$ time per iteration
because we have to assume the clusters to be unbalanced, and contain up to $O(n)$ objects;
making this $k$ times faster than regular PAM.
(The runtime of $O(nk)$ claimed by \citealp{DBLP:journals/eswa/ParkJ09}, clearly is incorrect.)

This heuristic is, unfortunately, not very effective at improving the clustering:
new medoids always \emph{have} to cover the entire current cluster.
This misses many improvements where cluster members can be reassigned to \emph{other} clusters
with little loss; such improvements are considered by SWAP.
Furthermore, the arithmetic mean as used in $k$-means changes when we move any
point to a different cluster, but the medoids will very often remain the same
even when objects are added to or removed from the cluster. 
Because of its \emph{discrete} nature, $k$-medoids is much more likely to get stuck
in a local optimum using this strategy.
Performing only \emph{few} swaps reduces the number of iterations (and hence reduces the run time),
but also produces significantly worse results,
as previously observed for example by \citet{DBLP:journals/ior/TeitzB68},
\citet{journals/geoana/Rosing79}, and \citet{DBLP:journals/jmma/ReynoldsRIR06}.
The problem of the $k$-means style ``alternating'' heuristic is illustrated in
\reffig{fig:alternating}.
Given the solution in the top row, assigning each object to the closest medoid
yields the indicated boxes as partitions; choosing the medoid (or median, as this
is a one-dimensional data set) in each box reproduces the same medoids---the
``alternating'' heuristic is stuck in a local optimum.
The SWAP heuristic can escape this situation easily:
swapping $A$ with medoid $M_1\!=\!B$ yields an improvement, because objects
$M_1\!=\!B$ and $C$ are reassigned to $M_2\!=\!E$ now.
In a second swap, $M_2\!=\!E$ will then be swapped with $D$.
The essential difference between the capabilities of the ``alternating'' and the SWAP heuristic
is that in the ``alternating'' heuristic, the new medoid \emph{must} cover all assigned points,
whereas when swapping medoids, points will be reassigned during the swap. Since SWAP can
reassign $B$ and $C$ to $M_2\!=\!E$ during the swap, it can choose a much better replacement
medoid than in the $k$-means style approach.
This example also can serve as a proof that the solutions found with this
heuristic can be arbitrarily much worse than the solution found by PAM,
if we move point $A$ further to the left.
(A similar situation can be found in Figure~1 of \citealp{DBLP:conf/nips/NewlingF17},
which serves as their motivation to use CLARANS for initializing $k$-means, and
a similar example is used by \citealp{DBLP:journals/mp/Hochbaum82} to discuss known
theoretical bounds on the quality: the longer the maximal edges in the graph are,
the worse approximations to the result become;
\citealp{DBLP:journals/comgeo/KanungoMNPSW04} used a similar example to show that
the worst case of the standard $k$-means algorithm is also not bounded, and propose
a PAM-style swapping approach for $k$-means to guarantee an approximation quality).

If we try to improve the ``alternating'' heuristic by allowing to reassign points to
other medoids, it essentially becomes a restricted variant of SWAP, where points may
only be swapped with the medoid they are currently assigned to. This yields little
benefit over the original SWAP, and even less over the accelerated version discussed
in \refsec{sec:fastpam1} where we can consider all medoids at once.

\subsection{Alternative Initializations}
While PAM's BUILD is considered a state of the art heuristic to initialize $k$-medoids
(it is also known by the name ``Greedy'' in operations research), some alternatives have
been used, such as randomly choosing initial medoids.

\citet{journals/ejor/Captivo91} integrates the ``Alternating'' approach into the
greedy BUILD heuristic, updating the medoid of the existing clusters
when the next center is added. This reportedly gives better starting conditions
than BUILD, but also requires more effort. This approach is known as ``GreedyG'',
and we will include it in our experiments.

\citet{DBLP:conf/soda/ArthurV07} noted that their $k$-means++ initialization heuristic
can also be used with linear error, and hence it can be used for $k$-medoids.
\citet{DBLP:conf/sisap/SchubertR19} experimented with this heuristic, but noted
on the benchmarks that it is very slow when used with PAM; we will investigate this below.

\cite{DBLP:journals/eswa/ParkJ09} propose an unusual $O(n^2)$ initialization,
that unfortunately tends to choose all initial medoids close to the center of the
data set. They choose the $k$ objects with the smallest normalized distance sums,
but ignore the dependency of the selected medoids to each other;
and hence this usually ends up choosing all medoids close to the 1-medoid.
This is not a very beneficial way of initializing these algorithms, and
indeed this tends to perform even worse than random sampling.
This was also previously observed by \citet{DBLP:conf/aistats/NewlingF17}.

\subsection{Variants for Large Data Sets}
Since PAM needs $O(n^2)$ memory for the distance matrix, it is not usable on big data.
Therefore, people have proposed various approximations to PAM, such as CLARA and CLARANS
discussed before.
\cite{serss/YangL14} parallelize the ``$k$-means like'' variant with map-reduce,
parallelizing over the cluster in the reduce step. When cluster sizes vary substantially,
this needs $O(n^2)$ memory in the reducer, and may yield next to no speedup in the worst case.
CLARA can be trivially parallelized by randomly partitioning the data,
then running PAM on each partition \citep{journals/compstat/KaufmanLR88}. %
This approach will obviously benefit from our improvements the same way as CLARA and PAM benefit.
A recent example is PAMAE~\citep{DBLP:conf/kdd/Song0H17}, which essentially is
CLARA with an additional refinement step: it draws random samples and runs
any $k$-medoids approach on each; chooses the best medoids found, and refines them
with a single iteration of an approximate parallel version of the ``$k$-means like''
update; this will benefit from our improvements to CLARA.
Papers have rarely considered using large $k$ values,
although this makes sense in the context
of approximating a big data set, where you want to reduce the data set to $k$ representative samples.
Many of the attempts at distributing and parallelizing PAM employ
PAM as a subroutine on a subset of the data, and hence can trivially integrate our improvements.

\section{Finding the Best Swap}\label{sec:best-swap}

We focus on improving the original PAM algorithm here, which is a commonly
used subroutine even in the faster variants such as CLARA.
We also discuss how we can obtain similar improvements for CLARANS in \refsec{sec:better-clarans}.

PAM's algorithm SWAP evaluates every possible swap of each medoid~$m_i$ with any non-medoid candidate~$x_c$.
Recomputing the resulting $\TD$ using \refeqn{eqn:td} every time would require finding the nearest medoid
for every point every time, which causes many redundant computations.
Instead, PAM computes the \emph{change} in $\TD$ for each object~$x_o$
if we swap $m_i$ with~$x_c$: %
\begin{align}
\Change =&
\textstyle\sum\nolimits_{x_o} \change(x_o,m_i,x_c)
\quad.
\label{eqn:td2}
\end{align}
In the function $\change(x_o,m_i,x_c)$ we can often detect when a point remains
assigned to its current medoid (if $\textit{nearest}(o)\!\neq\! i$, and this distance is also smaller than the distance to $x_c$),
and then immediately return $0$.
We rewrite the original ``if'' case distinctions
used in~\cite{KauRou90a} into the equation:
\begin{align}
\change(x_o,m_i,x_c)\!=\!&
\begin{cases}
0
&
\text{if }
\dist(x_o,x_c) \geq \dist_n(o) \text{ and } \textit{nearest}(o)\neq i
\quad\text{ (a)}
\\
\dist(x_o,x_c) - \dist_n(o)
&
\text{if }
\dist(x_o,x_c) < \dist_s(o) \text{ and } \textit{nearest}(o)= i
\quad\text{(b1)}
\\
\dist_s(o) - \dist_n(o)
&
\text{if }
\dist(x_o,x_c) \geq \dist_s(o) \text{ and } \textit{nearest}(o)=i
\quad\text{(b2)}
\\
\dist(x_o,x_c) - \dist_n(o)
&
\text{if }
\dist(x_o,x_c) < \dist_n(o) \text{ and } \textit{nearest}(o)\neq i
\quad\text{ (c)}
\end{cases}
\,.
\label{eqn:change}
\end{align}
\noindent
where $\dist_n(o)$ is the distance to the nearest medoid of $o$,
and $\dist_s(o)$ is the distance to the second nearest medoid.
The labels (a), (b1), (b2), and (c) indicate the if cases considered
by \citet{KauRou90a}.
If $\nearest(o)$, $\dist_n(o)$, and $\dist_s(o)$ are known,
we can compute $\change(x_o,m_i,x_c)$ using \refeqn{eqn:change} in $O(1)$,
and hence $\Change$ using \refeqn{eqn:td2} in $O(n-k)$ by skipping the selected medoids.
A naive approach would require $O(k)$ for $\change(x_o,m_i,x_c)$ respectively $O(nk)$ for computing $\Change$.

\cite{DBLP:journals/jmma/ReynoldsRIR06} note that we can decompose $\Change$ into:
(i)~the (positive) loss of removing medoid $m_i$, and assigning all of its members to the next best alternative,
which can be computed as $\Change^{-m_i}:=\sum_{\nearest(o)=i}\dist_s(o)-\dist_n(o)$, corresponding to case (b2) above,
and
(ii)~the (negative) loss of adding the replacement medoid $x_c$, and reassigning all objects closest
to this new medoid, computed as $\max\{\dist(x_o,x_c)-\dist_n^{-i}(o),0\}$,
where $\dist_n^{-i}(o)$ is the distance
to the nearest medoid except $m_i$ which we are replacing.
Since (i) as well as $\dist_n^{-i}(o)$ do not depend on the choice of $x_c$,
we can make the loop over all medoids $m_i$ outermost,
reassign all points of the current medoid to the second nearest medoid,
cache these distances to the now nearest neighbor as $\dist_n^{-i}(o)$,
and compute the resulting loss as:
\begin{align*}
\Change(m_i,x_c) := \Change^{-m_i} + \sum\nolimits_{x_o} \max\{\dist(x_o,x_c)-\dist_n^{-i}(o),0\}
\end{align*}
This combines cases (a) with (b2) and (b1) with (c),
moving the case distinction within these pairs of situations
outside of the innermost loop. These two cases still need to be distinguished in form of the $\max$ operation.
The authors observed roughly a two-fold speedup using this approach,
and so do we in our experiments.

The FastPAM idea is based on a similar idea of exploiting redundancy in these computations,
but we want to eliminate the nested loop over the medoids~$m_i$, hence removing the factor $k$
in the run time complexity.
This is possible because the four cases are not occurring equally often. No change is the
most common situation (at least for large $k$), and $\nearest(o)=i$ only holds for
roughly one of $k$ cases. The common cases (a) and (c) in \refeqn{eqn:change} do not depend
on the exact choice of $i$, as long as it is not the nearest medoid.
In fact, in a loop over all medoids $i$, all but one iteration perform the same computations.
When $\dist(x_o,x_c)<d_n(o)$ we get the same result independently of $i$ from cases (b1) and (c).
Hence we can rewrite this to:
\begin{align}
\change(x_o,m_i,x_c)\!=\!&
\begin{cases}
\dist(x_o,x_c) - \dist_n(o)
&
\text{if }
\dist(x_o,x_c) < \dist_n(o)
\\
\dist(x_o,x_c) - \dist_n(o)
&
\text{else if } \textit{nearest}(o)=i \text{ and } \dist(x_o,x_c) < \dist_s(o)
\\
\dist_s(o) - \dist_n(o)
&
\text{else if } \textit{nearest}(o)=i \text{ and } \dist(x_o,x_c) \geq \dist_s(o)
\\
0
&
\text{otherwise}
\end{cases}
\,.
\label{eqn:better-change}
\end{align}
In \citealt{DBLP:conf/sisap/SchubertR19} we still handled the first case with an
if-guarded nested loop over the medoids,
and argued that the loop is executed only for a subset of cases.
This does not allow for a worst-case guarantee, but based on the suggestion by Karl Bringmann
in personal communication we found a way to eliminate the nested loop altogether:
The first case can be efficiently handled by adding the change to a shared accumulator $\Change^{+x_c}$.
For the next two cases, we use an array with one entry for each medoid~$m_i$, and
if $\dist(x_o,x_c)\geq\dist_n(o)$, we collect the loss change for this one medoid~$m_i$ there separately.
The last case does not need to be handled, because it is~0. The final loss can then be obtained by adding the
shared accumulator $\Change^{+x_c}$ to all array entries (but only needed for the best).
Formally, this can be expressed as
\vspace{-\abovedisplayskip}
\begin{align}
\Change^{+x_c}=&
\sum_{x_o}
\kern10pt
\begin{cases}
\dist(x_o,x_c)-\dist_n(o)
&%
\text{if }\dist(x_o,x_c)<\dist_n(o)
\\
0
&%
\text{otherwise}
\end{cases}
\\
\Change(m_i,x_c)=
\Change^{+x_c}\mskip\thickmuskip{}+&
\sum_{\mathclap{\textit{nearest}(o)=i}}
\kern10pt
\begin{cases}
\dist(x_o,x_c)-\dist_n(o)
&%
\text{if }\dist(x_o,x_c)<\dist_s(o)
\\
\dist_s(o)-\dist_n(o)
&%
\text{otherwise}
\end{cases}
\label{eqn:better-change2}
\end{align}
which can be computed in a single pass over the objects.
But when benchmarking this method, we found that it was slower than that of
\citeauthor{DBLP:journals/jmma/ReynoldsRIR06} for $k\leq 3$, while for larger $k$ it
was much faster. Then we realized that we can integrate this idea, too. While
it remains slower for $k = 2$, it further increases the performance of the
algorithm; and $k = 2$ could become a special case in an optimized implementation.
Our final FastPAM SWAP proposal uses
\begin{align}
\Change^{-m_i}=&
\hphantom{+}
\sum_{\mathclap{\nearest(o)=i}}
\kern10pt
\dist_s(o)-\dist_n(o)
\\
\Change^{+x_c}=&
\hphantom{+}
\sum_{x_o}
\kern10pt
\begin{cases}
\dist(x_o,x_c)-\dist_n(o)
&%
\text{if }\dist(x_o,x_c)<\dist_n(o)
\\
0
&%
\text{otherwise}
\end{cases}
\\
\Change(m_i,x_c)=&
\Change^{-m_i}+
\Change^{+x_c}\notag\\&{+}
\sum_{\mathclap{\textit{nearest}(o)=i}}
\kern10pt
\begin{cases}
\dist_n(o)-\dist_s(o)
&%
\text{if }\dist(x_o,x_c)<\dist_n(o)
\\
\dist(x_o,x_c)-\dist_s(o)
&%
\text{else if }\dist(x_o,x_c)<\dist_s(o)
\\
0
&%
\text{otherwise}
\end{cases}
\label{eqn:better-change3}
\end{align}
where $\Change^{-m_i}$ is the same as in our presentation of
\citeauthor{DBLP:journals/jmma/ReynoldsRIR06},
except that we compute and store this for all $m_i$ in one pass, and do not use $d_n^{-m_i}\!$.
$\Change^{+x_c}$ is the same as just introduced. The last term---the only part which depends on
$x_o$---has become a more complicated expression, but it is frequently zero (and that is when
we save effort), and the first condition already needs to be evaluated for computing $\Change^{+x_c}$.
\\
In \ref{app:change-proof} we prove the relationship of
\refeqn{eqn:better-change3} to \refeqn{eqn:td2}.

In this article, we combine several optimization techniques:
\begin{enumerate}
\item[(A)] \emph{removal of the nested loop over the medoids}, the key contribution of the
initial FastPAM presented in \citet{DBLP:conf/sisap/SchubertR19}.
\item[(B)] \emph{introduction of the shared accumulator $\Change^{+x_c}$} (necessary to completely remove the loop,
in initial FastPAM it just became conditional)
\item[(C)] \emph{precomputation of removal loss} based on the idea of \cite{DBLP:journals/jmma/ReynoldsRIR06}
\item[(D)] \emph{eager execution of swaps}, going beyond the additional swaps we performed
in \citet{DBLP:conf/sisap/SchubertR19}, but similar to, e.g., CLARANS %
\end{enumerate}
With only techniques (A)-(C) implemented, we can still guarantee to find the exact same results
as the original PAM algorithm, we will denote this interim step as FastPAM1.
The full variant implementing (A)-(D) will be called FastEagerPAM, or short: ``FasterPAM''.
\citet{DBLP:conf/sisap/SchubertR19} used only~(A) and for FastPAM2 a simpler version of~(D)
that would perform up to $k$~swaps per iteration, one for each medoid.

\subsection{Making PAM SWAP Faster: FastPAM1}\label{sec:fastpam1}

\begin{algorithm2e*}[b!]\addtolength{\hsize}{1.5em}
\caption{FastPAM1: Improved SWAP algorithm}
\label{alg:iswap}
\SetKwFor{Repeat}{repeat}{}{end}
\SetKw{StopIf}{break loop if}
\lForEach{$x_o$}{
  compute $\nearest(o), \dist_{\nearest}(o), \dist_{\text{second}}(o)$\label{line:f1-nearest}%
}
\Repeat{}{
  $\Change^{-m_1},\ldots,\Change^{-m_k} \leftarrow $ compute removal loss\;
  $(\best{\Change}, \best{m}, \best{x}) \leftarrow (0, \text{null}, \text{null})$%
  \tcp*{Empty best candidate storage}\label{line:best}
  \ForEach(\tcp*[f]{Iterate over all non-medoids}){$x_c \not\in \{m_1,\ldots,m_k\}$}{
    $\Change_1,\ldots,\Change_k\leftarrow (\Change^{-m_1},\ldots,\Change^{-m_k})$\tcp*{Use removal loss}\label{line:swap1-init}
    $\Change^{+x_c}\leftarrow 0$\tcp*{Shared accumulator}
    \ForEach{$x_o$}{
      $d_{oj}\leftarrow \dist(x_o,x_c)$\tcp*{Distance to new medoid}
      \If(\tcp*[f]{\frenchspacing Case (i): nearest}){$\dist_{oj}<\dist_{\nearest}(o)$}{
        $\Change^{+x_c}\leftarrow\Change^{+x_c} + \dist_{oj} - \dist_{\nearest}(o)$\;\label{line:casea}
        $\Change_{\nearest(o)} \leftarrow \Change_{\nearest(o)}
           +\dist_{\nearest}(o) - \dist_{\text{second}}(o)$\;
      }\ElseIf(\tcp*[f]{\frenchspacing Case (ii): second nearest}){$\dist_{oj}<\dist_{\text{second}}(o)$}{
        $\Change_{\nearest(o)} \leftarrow \Change_{\nearest(o)}
           + \dist_{oj} - \dist_{\text{second}}(o)$\;\label{line:casebc}
      }
    }
    $i\leftarrow \argmin \Change_i$\tcp*{Choose best medoid i}
    $\Change_i \leftarrow \Change_i + \Change^{+x_c}$\tcp*{Add accumulator}
    \lIf(\tcp*[f]{Remember}){$\Change_i < \best{\Change}$}{
      $(\best{\Change}, \best{m}, \best{x}) \leftarrow (\Change_i, m_i, x_c) $
    }
  }
  \StopIf $\best{\Change}\geq 0$\;
  swap roles of medoid $\best{m}$ and non-medoid $\best{x}$\;
  \lForEach{$x_o$}{
    update $\nearest(o), \dist_{\nearest}(o), \dist_{\text{second}}(o)$\label{line:f1-update}%
  }
  $\TD \leftarrow \TD + \best{\Change}$\;
}
\Return $\TD,M,C$\;
\end{algorithm2e*}

\refalg{alg:iswap} shows the improved SWAP algorithm.
As we do \emph{not yet} decide which medoid to remove,
we use an array of $\Change_i$s for each possible medoid to replace,
initialized with the per-medoid removal loss $\Change^{-m_i}\!$.
Additionally, we employ a shared accumulator $\Change^{+x_c}$ to collect the loss change
independent of the medoid removed.
At the beginning of the iteration in line~\ref{line:swap1-init}, we use these precomputed
values as initial per-medoid accumulator values.
We can now iterate over all points, and check which of the three cases discussed above applies,
and the cases can be easily distinguished by using the
distance to the new candidate medoid $\dist(x_o,x_c)$, and the two cached distances
$\dist_{\nearest}(o)$, and $\dist_{\text{second}}(o)$.
If the new medoid is closest, the change applies to $\Change^{+x_c}$ (case (i), line~\ref{line:casea}),
but we have adjustments for the case that the nearest is removed (to not double-count this).
If the new medoid is second closest, we only have to handle the case where its current
nearest is removed (line~\ref{line:casebc}), and assign it to the new medoid then instead.
After iterating over all points we choose the best medoid, add the shared loss accumulator $\Change^{+x_c}\!$,
and remember the overall best swap.
Note that if we always prefer the smaller index $i$ on ties,
FastPAM1 carries out \emph{exactly the same} swap as the original PAM algorithm.

At the slight cost of precomputing the $k$ removal loss $\Change^{-m_i}$s,
temporarily storing the accumulator $\Change^{+x_c}$ and one $\Change_i$ for each medoid $m_i$
(compared to the cost of the distance matrix and the distances to the
nearest and second nearest medoids,
the cost of this is negligible),
we are able to remove the nested loop over all medoids,
hence making the PAM algorithm $O(k)$ faster.

\subsection{FasterPAM: FastPAM1 with Eager Swapping}\label{locality-gap}

Choosing the single best swap is not crucial to the optimization problem, and for example
``fast interchange'' of
\citet{journals/infor/Whitaker83} and CLARANS~\citep{DBLP:conf/vldb/NgH94,DBLP:journals/tkde/NgH02} are two approaches that greedily perform
the first swap that yields some improvement of the loss function.
FastPAM2 \citep{DBLP:conf/sisap/SchubertR19} was an interim solution: it first computed the
best swap for each medoid with a small modification to FastPAM1; then executed up to $k$
swaps per iteration (one for each medoid). In this section we will show that the simple eager approach is desirable to use.
\citet{DBLP:conf/pakdd/Estivill-CastroM98} point out that such ``local hill-climbing''
methods nevertheless guarantee to find a local optimum, but may provide much better runtime.
In facility location, both original PAM and FastPAM belong to the family of ``local search''
algorithms. A detailed analysis of result quality obtainable with such methods can be found,
for example, in \cite{DBLP:conf/stoc/AryaGKMP01,DBLP:journals/siamcomp/AryaGKMMP04},
who showed that the results obtained
by single-swap interchanges have a locality gap of exactly 5 as $k$ tends to infinity;
by considering multiple swaps this can be improved to a $3+\varepsilon$ bound at considerable
computational effort. The $k$-means++ algorithm is only $O(\log k)$ competitive \citep{DBLP:conf/soda/ArthurV07},
meaning that it can potentially yield much worse results than a swap heuristic
(c.f., \citealp{DBLP:journals/comgeo/KanungoMNPSW04}).

We can modify PAM as well as FastPAM to immediately perform any swap that yields
an improvement. We then continue searching at the next object, until we have
performed one entire pass over the data without finding an improvement.
Clearly this strategy can find many swaps per iteration.
Again the improvement of our modification saves a loop over the medoids,
meaning that FastEagerPAM is $O(k)$ faster than EagerPAM per iteration.
Because we compute all $k$ medoids in parallel, we can swap each candidate point with
the best of the $k$ medoids (if any yields a loss decrease).
We hence do not strictly perform the first swap, but the best of each batch of $k$.
This means this approach will find different results than both PAM and EagerPAM,
but there is no reason to assume that picking
the best of $k$ choices instead of the first yields worse results
(i.e., that the results
would be worse than those of eager swapping with regular PAM). It is less obvious that
this will yield as good results as a non-eager approach (PAM, respectively FastPAM1)
that considers the best swap only; the concern that we may get stuck in a worse local
optimum is larger when we do not try hard to choose the best swap. 
But if the solution space is fairly smooth -- and we would assume that the true best
solution is fairly well separated from inferior solutions if the clusters
are substantial and not random (this is a bit different from facility location,
where we are interested in finding optimal solutions even for data that does not
cluster; we may be interested to service all customers in the U.S.{} from
a given $k$~facilities even though the ``natural'' clustering would place one
facility in each larger city).

In \refalg{alg:eagerswap} we provide the pseudocode for this algorithm,
based on FastPAM1 (as we have only one candidate $x_c$ at any time).

\begin{algorithm2e*}[b!]\addtolength{\hsize}{1.5em}
\caption{FasterPAM: FastPAM1 with eager swapping}
\label{alg:eagerswap}
\SetKwFor{Repeat}{repeat}{}{end}
\SetKw{StopIf}{break outer loop if}
\SetKw{KwAnd}{and}
\SetKw{Where}{where}
$x_{\text{last}}\leftarrow \text{invalid}$\;
\lForEach{$x_o$}{
  compute $\nearest(o), \dist_{\nearest}(o), \dist_{\text{second}}(o)$
}
$\Change^{-m_1},\ldots,\Change^{-m_k} \leftarrow $ compute initial removal loss\;
\Repeat{}{
  \ForEach(\tcp*[f]{Iterate over all non-medoids}){$x_c \not\in \{m_1,\ldots,m_k\}$}{
    \StopIf $x_c = x_{\text{last}}$ \tcp*{No improvements found}
    $\Change\leftarrow (\Change^{-m_1},\ldots,\Change^{-m_k})$\tcp*{Use removal loss}
    $\Change^{+x_c}\leftarrow 0$\tcp*{Shared accumulator}
    \ForEach{$x_o$}{
      $d_{oj}\leftarrow \dist(x_o,x_c)$\tcp*{Distance to new medoid}
      \If(\tcp*[f]{Case (i)}){$\dist_{oj}<\dist_{\nearest}(o)$}{
        $\Change^{+x_c}\leftarrow\Change^{+x_c} + \dist_{oj} - \dist_{\nearest}(o)$\;
        $\Change^{+\nearest(o)} \leftarrow \Change^{+\nearest(o)}
           +\dist_{\nearest}(o) - \dist_{\text{second}}(o)$\;
      }\ElseIf(\tcp*[f]{Case (ii) and (iii)}){$\dist_{oj}<\dist_{\text{second}}(o)$}{
        $\Change^{+\nearest(o)} \leftarrow \Change^{+\nearest(o)}
           + \dist_{oj} - \dist_{\text{second}}(o)$\;
      }
    }
    $i\leftarrow \argmin \Change_i$\tcp*{Choose best medoid}
    $\Change_i \leftarrow \Change_i + \Change^{+x_c}$\tcp*{Add accumulator\ \ \ }
    \If(\tcp*[f]{Eager swapping\ \ \ \ }){$\Change_i < 0$}{
      swap roles of medoid $\best{m}$ and non-medoid $x_o$\;
      $\TD \leftarrow \TD + \Change_i$\;
      update $\Change^{-m_1},\ldots,\Change^{-m_k}$\;
      $x_{\text{last}}\leftarrow x_o$\;
    }
  }
}
\Return $\TD,M,C$\;
\end{algorithm2e*}

\subsection{Faster Initialization}\label{sec:fasterinit}

With these optimizations to the PAM~SWAP algorithm,
reducing the run time from \mbox{$O(k(n-k)^2)$} to \mbox{$O((n-k)n)$},
the bottleneck of PAM becomes the BUILD phase.
In the experiments with the 100 plant species leaves data sets in \refsec{sec:experiments}
with $k = 100$, PAM spends 95\% of the run time in SWAP.
With FastPAM1, this reduced to about 19\%; and with FasterPAM only 3.7\% are spent in SWAP.
Matrix computation takes 0.24\%, 3.6\%, respectively 4.1\% of the total runtime.
The amount of time spent in BUILD however increases from 5.3\% to 76\% respectively 91\%.
Since the complexity of BUILD is in $O(kn^2)$, this should not come at a surprise that it now
has become the dominant bottleneck in the algorithm.
But because we made SWAP much faster, we can afford to begin with
slightly worse starting conditions, although this means we need more iterations of SWAP afterwards
(in the experiments of \refsec{sec:experiments}, we will see that we need to do many more swaps
with worse starting conditions, and that BUILD was indeed beneficial to use with the original PAM).

An elegant way of initializing $k$-means is $k$-means++~\citep{DBLP:conf/soda/ArthurV07}.
The beautiful idea of this approach is to choose seeds with the probability proportional
to their squared distance to the nearest seed (the first seed is picked uniformly).
While this approach is commonly known as $k$-means++, an earlier version and analysis
of this idea can already be found in \cite{DBLP:conf/focs/Meyerson01} in the context
of the online facility location problem, while \cite{DBLP:conf/focs/OstrovskyRSS06} published
a variant that also uses importance weighting for the first seed.
If we assume there exists a cluster of several points and no seed nearby,
the aggregated probability mass of this cluster is substantial,
and we are likely to place a seed there; afterwards the probability
mass of this cluster reduces and we are unlikely to place a second seed there.
Outliers on the other hand will have a high individual weight,
but as they are rare their total mass remains low enough to usually not be chosen. 
This initialization is (in expectation) $O(\log k)$ competitive to the optimal solution,
so it will theoretically generate better starting conditions than uniform random sampling.
But as seen in our experiments, this guarantee is pretty loose; and BUILD empirically
produces much better starting conditions than $k$-means++
(although theoretical bounds known due to
\citealp*{journals/mnsc/CornuejolsFN77,DBLP:journals/mp/Hochbaum82} are not encouraging).
But it is easy to see that in BUILD each medoid is chosen as a current optimum with respect to \TD{};
whereas $k$-means++ picks the first point randomly, and subsequent points are (in expectation)
\emph{random} points from different clusters, but $k$-means++ makes no effort to find
\emph{good} centers of the clusters
(which is not that important for $k$-means, where the mean is in between of the data points).
We use the term ``distance weighted'' initialization in the following for the canonical adaption of the $k$-means++
initialization strategy from squared Euclidean distances to arbitrary distances.
Therefore, with distance weighed initialization we need around $k$ additional swaps
to pick the medoid of each cluster (and hence, $k$~SWAP iterations of original PAM and FastPAM1).
Because a single iteration of SWAP used to take as much time as BUILD,
the distance weighed initialization only begins to shine if we use
FastPAM1 to reduce the cost of iterating together with the eager swapping
of FasterPAM doing as many swaps as possible in each iteration.
A second important benefit of $k$-means++ is that the algorithm is randomized;
and we can run it multiple times and keep the best result.
This helps if there is some local optimum that we might get caught in, and we
can use multiple runs to increase our likelihood of finding the true optimum.
\cite{DBLP:journals/datamine/LijffijtPP15} previously used k-means++ for PAM and CLARA;
but they mistake the ``alternating'' algorithm for PAM, and their experiments only
used small~$k$. Our experiments (in \refsec{sec:numiter}) show that
distance weighed initialization takes \emph{many more iterations} to converge
than with the original BUILD initialization;
so without the improvements introduced in this
article, it is usually not beneficial to use distance weighed initialization with the original SWAP algorithm for speed
(the benefit of randomness, the ability to get different results, remains;
and so do the theoretical guarantees).

\citet{DBLP:conf/sisap/SchubertR19} proposed a strategy called LAB (Linear Approximative BUILD),
a linear approximation of the original PAM BUILD (c.f., \refalg{alg:build}).
In order to achieve runtime linear in $n$, we simply subsample the data set.
Before choosing each medoid, we sample $10+\lceil\!\sqrt{n}\;\rceil$ points from all non-medoid points.
From this subsample we choose the one with the largest decrease $\Change$ with respect to the
current subsample only, similar to BUILD.
Other similar sampling-based heuristics have been used, for example, by \citet{journals/heuristic/ResendeW04},
whose strategy yields a complexity of $O(kn\log_2 \tfrac{n}{k})$, and this approach
performed best in their experiments.
While LAB reduced the number of swaps necessary for convergence (and hence the number of iterations
for PAM and FastPAM1) substantially, this benefit was offset when we introduced eager swapping
in FasterPAM. With FasterPAM, we recommend using either uniform random sampling or distance weighed
initialization.

\subsection{Integration: FastCLARA and FastCLARANS}
\label{sec:better-clarans}

Since CLARA~\citep{KaufmanR86,KauRou90b}
uses PAM as a subroutine, we can trivially
use our improved FastPAM with CLARA. In the experiments
(the implementations are provided as open-source)
we will denote this variant as FastCLARA.

Because of the sampling strategy used in CLARA,
the runtime is $O(k^3\cdot i)$ because it performs PAM clustering
with a sample size of $s\!=\!40+2k$ (the number of iterations~$i$ does, however,
contain some hidden dependency on~$k$).
By replacing PAM with FastPAM, we immediately obtain a runtime of $O(k^2\cdot i)$
for CLARA this way.
With modern hardware we do, however, suggest to also use
at least twice as many samples as in the original recommendation,
i.e., $s\!=\!80+4k$.
By choosing a sample size in $O(\!\smash{\sqrt{n}})$, we can also obtain a
$O(n\cdot i)$ time approximation to PAM, for example we may choose to
use $s\!=\!\smash{\sqrt{n}}+4k$.
While we must assume that the worst case for $i$ is, unfortunately, similar to $k$-means and hence in the order of
$\smash{i\!\in\!O(2^{\!\sqrt{s}})}$ for sample size~$s$ (c.f., \citealp{DBLP:conf/compgeom/ArthurV06}),
this will give decent results in seemingly linear time for many practical purposes.

CLARANS~\citep{DBLP:conf/vldb/NgH94,DBLP:journals/tkde/NgH02} uses a randomized search instead of considering all possible swaps.
For this, it chooses a random pair of a non-medoid object and a medoid, computes whether this
improves the current loss, and then eagerly performs this swap.
Adapting the idea from FastPAM1 to the random exploration approach of CLARANS,
we pick only the non-medoid object at random,
but can consider all medoids at a similar run time to looking at a single medoid.
This means we can either explore $k$ times as many edges of the graph,
or we can reduce the number of samples to draw by a factor of $k$.
In our experiments we opted for the second choice (sampling $2.5\%\cdot (n-k)$
non-medoids, rather than $1.25\%\cdot k\cdot(n-k)$ edges), to make the results
scale similar to the original CLARANS. %
By varying the subsampling rate, the user can control the tradeoff
between computation time and exploration.

\section{Experiments}\label{sec:experiments}

By eliminating the nested loop over the medoids, we must expect an $O(k)$ speedup
of FastPAM1 over the original PAM algorithm (in contrast to much work published
in recent years, the speedup is not just empirical).
Nevertheless constant factors and implementation details can make a
big difference~\citep{DBLP:journals/kais/KriegelSZ17}, and we want to ensure that
we do not pay big overheads for theoretical gains that would only manifest
for infinite data.\footnote{Clearly, our $O(k)$ fold speedup must be immediately
measurable, not just asymptotically, because the constant overhead for maintaining
the fixed array cache is small.}
Because of constant factors, it could for example be possible that we need a
certain minimum $k$ for this approach to be beneficial over the original PAM.
For the additional benefit of eager swapping in FastEagerPAM we do not have
theoretical guarantees for an additional speedup over FastPAM1;
the resulting speedup is expected to be a much smaller
factor due to the reduction in iterations, at the price of performing more swaps.
In contrast to FastPAM1, these do not guarantee the identical results as original PAM;
therefore we also want to verify that they are of the expected equivalent quality.
But because we observed the initialization time to become a bottleneck now,
and we also propose to use different initialization techniques,
we will first of all evaluate picking the initial medoids; then proceed
with the experiments regarding the scalability in~$k$ and~$n$
as well as result quality.

As discussed before, all these algorithms belong to the family of local search
optimization algorithms.
As long as there is a reachable state that yields an improvement, this change is applied.
The approaches considered in our experiments differ by
(i)~which local changes are considered, and
(ii)~whether only the best change is applied, or the first.
Because of the local search, these algorithms can get stuck in local minima;
where one would expect that (i)~has much more effect on the result quality,
whereas (ii)~primarily affects performance.

\subsection{Research Questions}

We conceptualized our experiments to verify the performance and quality
of our new method compared to previous work.
In particular, we aim at empirically answering the following research questions:

\begin{enumerate}
\item
How do the different initialization methods affect result quality?
\item
How do the different initialization methods affect run time of the algorithms?
\item
Are there particularly favorable combinations of algorithms and initializations?
\item
What is the practical speedup over the original PAM algorithm
in particular in dependence on the parameter $k$ (known to be $O(k)$ from theory)?
\item
How many iterations can eager swapping save over optimal swapping, and how many more
swaps does it perform until convergence?
\item
How does the result quality compare to approximative algorithms?
\item
Can results be replicated on a different data set?
\item
How do the algorithms scale with the data set size?
\end{enumerate}

The first three research questions will be studied in \refsec{sec:init} focused on initialization.
Research question four will be evaluated in \refsec{sec:speedupk}, question five in \refsec{sec:numiter}.
Approximative algorithms are compared in \refsec{sec:quality}.
In \refsec{sec:optdigits} we replicate the results on an additional data set.
Scalability with data set size is evaluated in \refsec{sec:scalability} on another, larger, data set.
Before beginning with the experiments, we first describe the data sets and test environment in the next section.

\subsection{Data Sets}

In order to test the quality of the different initialization methods and approaches,
we use a classic library of 40 $k$-median problems from operations research.
These can be found in the OR-Library\footnote{Data available at
\url{http://people.brunel.ac.uk/~mastjjb/jeb/info.html}}
\citep{journals/jors/Beasley90},
which is known for its traveling salesman problem sets.
By today's standards these problems are fairly small (up to 900 instances),
\emph{but} for these problems the true optimum solution has been
determined. Hence, we can compute the gap between the solution found by the algorithm
and the true optimum solution on some well-known example problems. In \refsec{locality-gap}
we discussed that a local minimum can be up to 5 times larger than the optimum,
but we will show here that the difference usually is much smaller, making these approximations
sufficient for many applications.
In addition to the ORlib problems, we also consider three variants of problems from this data set first used by
\citet{book/LagunaV90/SenneL}: sl700, sl800, and sl900 are problems pmed34, pmed37, and pmed40
respectively, but with a larger $k\!>\!200$ (making our improvements even more effective
for these problems). The data sets gr100 and gr150 are graph data sets that originate from
\citet{journals/ejor/GalvaoR96}, and have been used by \cite{book/LagunaV90/SenneL}
for benchmarking $k$-medoids. There are true optimum results given for $k = 5,10,15,20,25,30,40,50$
by \citet{journals/heuristic/ResendeW04},\footnote{Data available at
\url{http://mauricio.resende.info/popstar/downloads.html}}
which we use for evaluation.

To test scalability we need larger data sets than these classic problems.
We will be using three data sets from the well-known UCI repository \citep{Dua:2019}:
First we will showcase results using the texture features from the
``one-hundred plant species leaves'' data set, which we chose
because it has 100 classes, and 1600 instances,
a fairly small size that regular PAM can still easily handle.
Naively, one would expect that $k = 100$ is a good choice on this data set,
but some leaf species are likely not distinguishable by unsupervised learning.
The same experiment is repeated on the ``Optical Recognition of Handwritten Digits''
data set with $n\!=\!5620$ instances, $d\!=\!64$ variables,
and $10$ natural classes and the well-known MNIST data set,
which has 784 variables (each corresponding to a pixel in a $28\!\times\!28$ grid)
and 60000 instances (PAM will not be able to handle this size in reasonable time
anymore).

We used the ELKI open-source data mining toolkit~\citep{DBLP:journals/corr/abs-1902-03616}
in Java to develop our version.
In prior work \citep{DBLP:conf/sisap/SchubertR19},
we also reproduced the results with the R \texttt{cluster} package,
which is based on the original PAM source code and written in C.
We omit results of the R version for redundancy in this article.
The experiments were processed using server
Intel Xeon E5-2697v2 CPUs at 2.70GHz in a small cluster.
There is an additional (but not significant) speedup possible by
closer integration of the initialization with the algorithm in some cases,
where the closest and second medoid are already computed in the initialization,
and hence line~\ref{line:swap-nearest} in SWAP (\refalg{alg:swap}, similar in
the other algorithms) could be avoided for \emph{some} initialization methods
(but not for uniform random initialization, obviously).
For the special case of $k=2$, several additional optimizations are possible,
as the second nearest medoid must always be the only other medoid; we also
did not want to over-optimize for this case in our experiments,
although it may arise in many practical applications.

\subsection{Initialization}\label{sec:init}

\begin{table}\centering\small\setlength{\tabcolsep}{3.5pt}
\begin{tabular}{l|rrrrrrrr}
Initialization & mean    & min     & optimal & $\sigma_{\text{rand}}$ & $\sigma_{\text{data}}$ & time    & iterations & swaps  
\\\hline
              &\multicolumn{8}{c}{\bf Initialization}\\\hline
BUILD         &    4.0\% &    3.7\% &        4 &    0.2\% &    3.0\% &    157 ms &          &          \\
GreedyG       &    2.7\% &    2.4\% &        6 &    0.3\% &    2.9\% &    206 ms &          &          \\
LAB           &   37.6\% &   27.3\% &        0 &    6.8\% &   12.6\% &      9 ms &          &          \\
distance weighted &   94.9\% &   71.2\% &        0 &   15.5\% &   15.1\% &      4 ms &          &          \\
Random        &   99.6\% &   74.6\% &        0 &   16.8\% &    6.2\% &      0 ms &          &          \\
Park and Jun  &  115.7\% &  115.7\% &        0 &    0.0\% &   60.9\% &     14 ms &          &          \\
              &\multicolumn{8}{c}{\bf PAM}\\\hline
BUILD         &    1.2\% &    1.0\% &       22 &    0.2\% &    1.9\% &   1101 ms &      8.2 &      7.2 \\
GreedyG       &    1.2\% &    0.8\% &       25 &    0.2\% &    1.9\% &    659 ms &      5.0 &      4.0 \\
LAB           &    1.7\% &    0.5\% &       28 &    1.0\% &    2.4\% &   3671 ms &     28.1 &     27.1 \\
distance weighted &    1.7\% &    0.6\% &       31 &    0.9\% &    2.5\% &   5067 ms &     36.6 &     35.6 \\
Random        &    1.6\% &    0.4\% &       33 &    0.9\% &    2.4\% &   5271 ms &     38.4 &     37.4 \\
Park and Jun  &    1.2\% &    1.0\% &       23 &    0.2\% &    1.9\% &   6298 ms &     42.6 &     41.6 \\
              &\multicolumn{8}{c}{\bf FastPAM1}\\\hline
BUILD         &    1.2\% &    1.0\% &       22 &    0.2\% &    1.9\% &    184 ms &      8.2 &      7.2 \\
GreedyG       &    1.2\% &    0.8\% &       25 &    0.2\% &    1.9\% &    232 ms &      5.0 &      4.0 \\
LAB           &    1.7\% &    0.5\% &       28 &    1.0\% &    2.4\% &     75 ms &     28.1 &     27.1 \\
distance weighted &    1.7\% &    0.6\% &       31 &    0.9\% &    2.5\% &     78 ms &     36.6 &     35.6 \\
Random        &    1.6\% &    0.4\% &       33 &    0.9\% &    2.4\% &     78 ms &     38.4 &     37.4 \\
Park and Jun  &    1.2\% &    1.0\% &       23 &    0.2\% &    1.9\% &     96 ms &     42.6 &     41.6 \\
              &\multicolumn{8}{c}{\bf EagerPAM}\\\hline
BUILD         &    1.2\% &    0.8\% &       23 &    0.3\% &    1.8\% &    300 ms &      2.5 &     11.0 \\
GreedyG       &    1.1\% &    0.7\% &       24 &    0.3\% &    1.8\% &    338 ms &      2.3 &      5.9 \\
LAB           &    1.5\% &    0.4\% &       30 &    0.9\% &    2.2\% &    244 ms &      3.6 &     91.4 \\
distance weighted &    1.4\% &    0.4\% &       31 &    0.9\% &    2.0\% &    312 ms &      4.1 &    168.5 \\
Random        &    1.4\% &    0.4\% &       31 &    0.8\% &    2.3\% &    339 ms &      4.3 &    194.7 \\
Park and Jun  &    1.2\% &    0.3\% &       33 &    0.7\% &    1.6\% &    420 ms &      4.4 &    276.7 \\
              &\multicolumn{8}{c}{\bf FasterPAM}\\\hline
BUILD         &    1.2\% &    0.8\% &       23 &    0.3\% &    1.8\% &    176 ms &      2.5 &     10.1 \\
GreedyG       &    1.1\% &    0.7\% &       24 &    0.3\% &    1.8\% &    226 ms &      2.3 &      5.6 \\
LAB           &    1.7\% &    0.5\% &       27 &    1.0\% &    2.2\% &     38 ms &      3.1 &     51.0 \\
distance weighted &    1.6\% &    0.5\% &       30 &    0.9\% &    2.2\% &     36 ms &      3.2 &     71.6 \\
Random        &    1.5\% &    0.4\% &       32 &    1.0\% &    2.1\% &     34 ms &      3.2 &     75.3 \\
Park and Jun  &    1.4\% &    0.4\% &       32 &    0.8\% &    1.7\% &     46 ms &      3.1 &     86.6 \\
              &\multicolumn{8}{c}{\bf Alternating}\\\hline
BUILD         &    3.4\% &    3.1\% &        4 &    0.2\% &    3.0\% &    185 ms &      1.5 &      0.5 \\
GreedyG       &    2.7\% &    2.3\% &        6 &    0.3\% &    2.9\% &    236 ms &      1.1 &      0.1 \\
LAB           &   27.0\% &   17.5\% &        2 &    6.4\% &   12.6\% &     47 ms &      2.4 &      1.4 \\
distance weighted &   49.3\% &   33.6\% &        0 &   10.4\% &   19.4\% &     44 ms &      3.0 &      2.0 \\
Random        &   55.9\% &   39.7\% &        0 &   10.9\% &   21.7\% &     42 ms &      3.1 &      2.1 \\
Park and Jun  &   69.7\% &   69.3\% &        0 &    0.4\% &   39.6\% &     51 ms &      3.4 &      2.4 \\
\end{tabular}
\caption{Comparison of different initialization procedures on the extended ORlib data sets.\newline
Scores are normalized to the known optimum solution (0\%) and the
average of 100 random medoids~(100\%).
``Mean'' and ``min'' are the average and best result over all data sets.
``Optimal'' denotes how many problems were solved optimally.
$\sigma_{\text{rand}}$ is the average standard deviation over 10 random restarts each;
$\sigma_{\text{data}}$ is the standard deviation of the mean over all data sets.
Time and number iterations are averaged over all restarts and data sets.}
\label{tab:init}
\end{table}

In \reftab{tab:init} we compare the quality of different initializations,
and how they affect the outcome of different algorithms.
As data sets we use the ORlib problems and the SL variants thereof,
along with the Galv\~ao and ReVelle graph data sets
because the optimum result is known for these problems.
For our quality measure we use a normalization inspired by \citet{journals/ejor/Captivo91},
\begin{align}
\text{Normalized Loss} &:=
(\TD-\TD_{\text{optimum}})/(\TD_{\text{random}}-\TD_{\text{optimum}}) \enskip, \label{eqn:norm}
\end{align}
because the simpler approach $\TD/\TD_{\text{optimum}}$ can be trivially
``improved'' by adding a constant to all non-zero distances.
A score of 0\% is obtained by the optimum solution, while picking medoids
uniformly at random yields an error close to 100\%.
We estimate $\TD_{\text{random}}$ using 100 random sampled medoids.
For methods that involve randomness, we report the average over 10 random restarts
and permutations of the data set,
and we additionally average over the 59 data sets. We always report the
standard deviation over the different data sets; for methods with randomness
we also report the average standard deviation over the restarts.
Additionally, we report in how many of the 59 problems the optimum solution could
be found with 10 restarts. Because some methods can swap multiple medoids in each
iteration, we list both the number of iterations and the number of swaps performed.

If we first look at the initialization in isolation,
GreedyG is the clear winner, closely followed by
BUILD. This is to be expected, because both try to optimize the objective
function \TD{} directly, and GreedyG includes an additional refinement step.
BUILD and GreedyG were able to find the optimum solution for $4$ resp.\ $6$ of the problems.
These two are, however, also the most costly.
While they are deterministic methods, the small variance observed is due to ties
when we shuffle the input data.
LAB offers a good performance, at a substantial quality improvement over the others.
Distance weighted initialization is already rather close to random in quality (this is
expected, because it attempts to distribute the centers, not to identify the
most central object of each partition).
Random, by definition, must score close to 100\% on this measure.
The initialization of \cite{DBLP:journals/eswa/ParkJ09} performs even worse
than random, because all medoids are placed very close to each other in the
center of the data set. It also shows a high standard deviation across data sets,
and would often perform twice as bad as random medoids.

If we now look at the result after running PAM,
we make the unexpected observation that a bad initialization does not have much
effect on the result quality of PAM; but it primarily affects run time.
By swapping, bad initial medoids can easily be replaced with better alternatives.
GreedyG still yields the best results on average, and despite being the slowest
initialization, it yields the fastest runtime for PAM. This is due to reducing
the number of swaps necessary for convergence significantly. The initialization
of Park scores among the best in average quality (despite the bad starting points),
but offers the worst runtime because of the high number of iterations.
Surprisingly slow with PAM is distance weighted initialization (also replicated on a second system),
supposedly because it is more likely to pick far points; and indeed the runtime
behavior closely matches that of a farthest-points initialization (not included in the table).
So while even bad initialization still allows us to find good results, it has a
major impact on runtime. Methods such as BUILD and GreedyG reduce the runtime
of PAM significantly, whereas heuristics that picking too-far or too-central
points can be worse than random. For PAM, the number of iterations is exactly
the number of swaps performed plus one, which is the final iteration where no
improving swap can be found anymore.
Considering PAM with BUILD and GreedyG, both of which are deterministic methods,
we nevertheless observe a standard deviation of 0.2\% depending on the input order
and duplicate distances. These 0.2\% establish a baseline that we must consider
to be random deviations when interpreting the entire table.

In $k$-means clustering experience has shown that multiple restarts can be
beneficial; and the random-based initializations (uniform random, distance weighted,
and LAB because of the sampling) can find better or worse results. It is worth
noting that with just 10 random restarts and keeping the result with lowest \TD{}
(which is our unsupervised optimization criterion) we can get significantly
better results with these three approaches than the averages reported in the
first column. As shown in the column titled ``optimal'', beginning with random centers
and 10 restarts was able to find the optimum solution on 33 of the data sets,
while the deterministic GreedyG initialization could only solve 25.
Unfortunately, the runtime with PAM and random initialization is several times higher,
and we may hence not be able to afford many restarts with PAM.

But there is no good reason to use the original PAM anymore -- FastPAM1 will find
the exact same results faster, and this experiment illustrates both that the results are
the same, and that we see substantial speedups.
The runtime includes initialization, there is some overhead included,
and it is aggregated over data sets of different size, difficulty, and different $k$
so we cannot expect the results to be exactly $k$ times faster.
But GreedyG now becomes the slowest method -- the runtime for this combination is now dominated by the initialization cost.
Because FastPAM1 spends 89\% of the time in initialization with GreedyG;
the overall speedup is only about $2.8\times$.
With PAM BUILD, the speedup is about $6\times$, and with random initialization,
where almost the entire time is spent in optimization, FastPAM1 was $67\times$ faster.
The average $k$ of the problems in this experiment is about 51 (when weighted with $n^2$,
the average $k$ increases to 83, as larger problems tend to have larger $k$).

As explained before, we expect sub-optimal swaps to not negatively affect quality;
and while we can save the time searching for a better swap, we have to pay the price
of performing maybe unnecessary swaps and updating our data structures.
For this, we investigate eagerly performing swaps; either based on the original
PAM algorithm (iterating over existing medoids, then non-medoids) denoted as EagerPAM,
or based on the FastPAM variant (iterating over non-medoids; combined computation of all medoids).
These two no longer find the exact same result, because EagerPAM iterates over medoids
in the outer loop, while FasterPAM only iterates over candidates and then considers the best
medoid for swapping only. Interestingly, while FasterPAM is able to choose better swaps
(the best of $k$; and hence it needs fewer swaps and iterations),
EagerPAM appears to find slightly better results than the others.
But these differences are within the 0.2\% margin of error that we attribute to
different processing order, and on other data sets
the FasterPAM approach also sometimes produces better results.
On the ORLib data, FasterPAM ran about twice as fast as FastPAM1.

It is remarkable that the random initializations like LAB, distance weighted, and even
uniform random become attractive now, because by performing multiple swaps per
iteration, bad starting conditions do not affect performance that much anymore.
Much of the drawbacks of distance weighted initialization observed in our prior
work \citep{DBLP:conf/sisap/SchubertR19} have now disappeared with FasterPAM,
where the average number of iterations is now similar to our proposed LAB initialization.
The number of swaps performed to convergence still differs substantially between good
and bad starting conditions, but this has much less impact on the number of iterations
or the overall runtime now. In this experiment, we cannot identify a clear winner
between LAB initialization (the slowest initialization, requiring the fewest iterations),
distance weighted, and uniform random initialization; this is largely because
the performance of the swapping procedure has become so good that the differences
disappear in the measurement uncertainty. Because of this, and because the performance
seems to be largely independent of the starting conditions, it seems adequate to prefer
the simplest initialization: uniform random sampling; and rather use multiple restarts
than computing better starting positions.
But we will study starting conditions again for example in~\reftab{tab:100plants}.

Last we want to look at the ``Alternating'' algorithm.
This is also a fast approach, but with a hefty quality decrease.
A good initialization with GreedyG and BUILD becomes very important.
With random initialization, the performance would usually be only about half-way
between the optimum and a completely random result; and with the problematic initialization of
\citet{DBLP:journals/eswa/ParkJ09} with 69.7\% much closer to random than to
the optimum. Even random initializations with 10 restarts do not help much here,
with the average best result still being 39.7\% worse than the optimum
(compared to values of about 0.4\% with 10 restarts of swap-based approaches).
Hence we advise against using the $k$-means style ``Alternating''
approach for \kmedoids{} because of result \emph{quality}
(contrary to the claims of \citealt{DBLP:journals/eswa/ParkJ09}, who only used very
simple data sets; but in line with earlier observations by, e.g., \citealt{DBLP:journals/ior/TeitzB68},
\citealt{journals/geoana/Rosing79}, and \citealt{DBLP:journals/jmma/ReynoldsRIR06}).

\subsection{Run Time Speedup with Increasing $k$}\label{sec:speedupk}

\begin{figure*}[tp!]
\begin{subfigure}{\linewidth}\centering
\includegraphics[width=.9\linewidth]{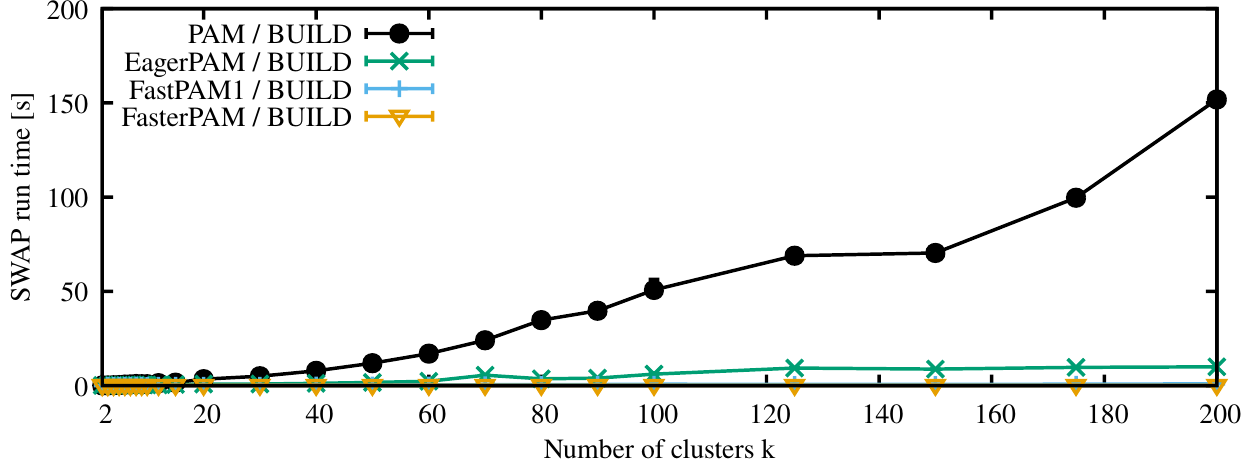}%
\vspace*{-.5ex}%
\caption{Run time in linear space}
\label{fig:100plants-opttime-lin}
\end{subfigure}
\\
\begin{subfigure}{\linewidth}\centering
\includegraphics[width=.9\linewidth]{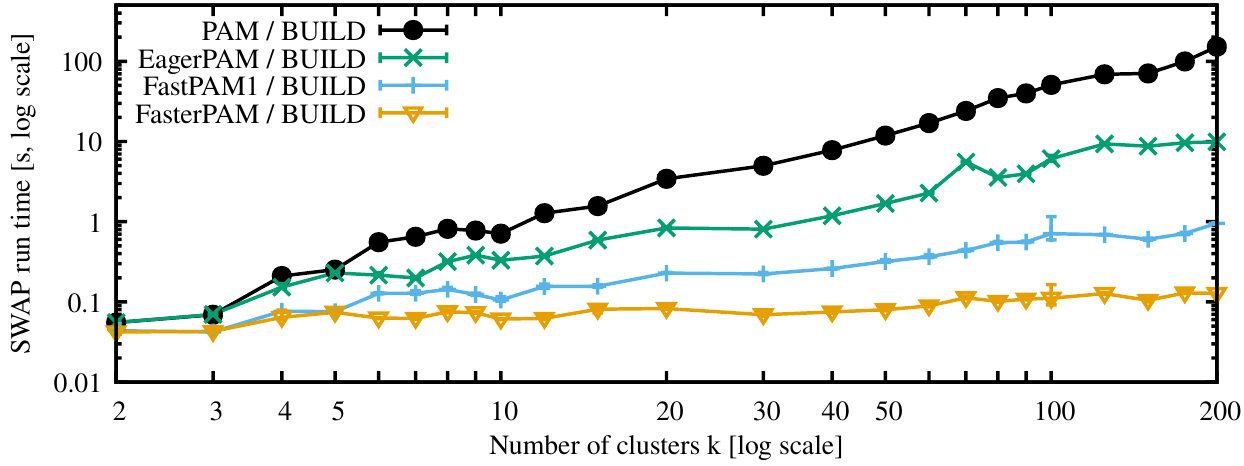}%
\vspace*{-.5ex}%
\caption{Run time in log-log space}
\label{fig:100plants-opttime-log}
\end{subfigure}
\\
\begin{subfigure}{\linewidth}\centering
\includegraphics[width=.9\linewidth]{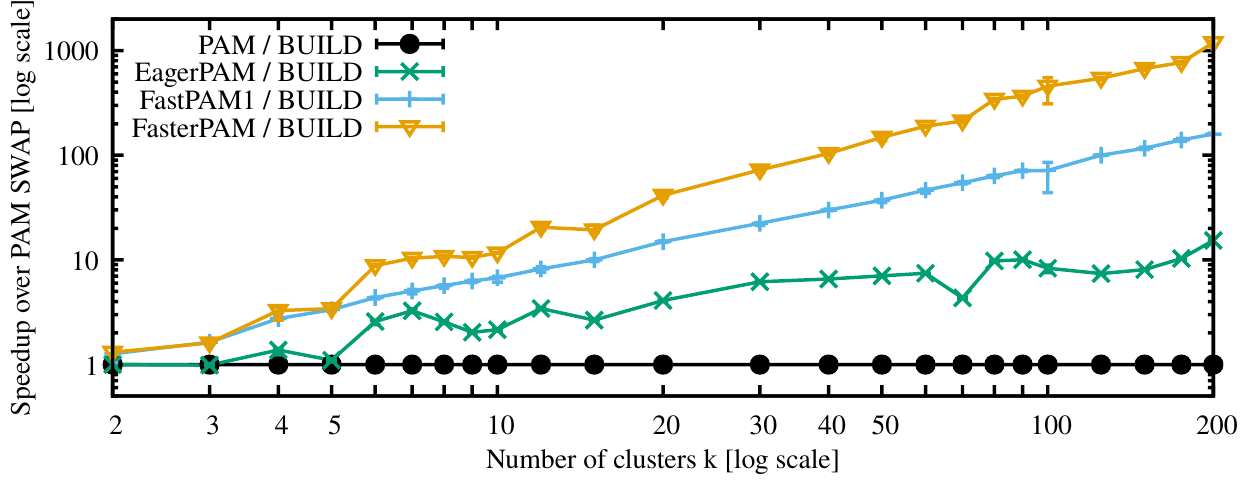}%
\vspace*{-.5ex}%
\caption{Speedup in log-log space}
\label{fig:100plants-optspeedup-log}
\end{subfigure}
\\
\begin{subfigure}{\linewidth}\centering
\includegraphics[width=.9\linewidth]{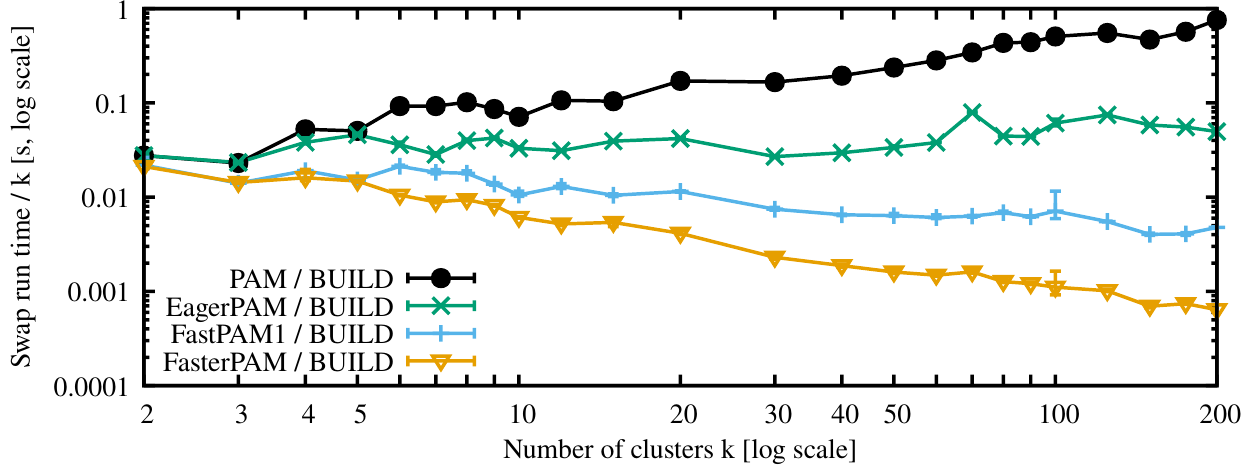}%
\vspace*{-.5ex}%
\caption{Runtime divided by the number of clusters $k$ [log scale]}
\label{fig:100plants-opttime-norm-log}
\end{subfigure}
\caption{Run time of PAM SWAP (SWAP only, without DAISY, without BUILD)}
\label{fig:100plants}
\end{figure*}

Next we want to explore the speedup as we increase $k$ on a data set.
While the theoretical speedup of both FastPAM and FasterPAM is $O(k)$,
this is only beneficial if it is measurable for realistic values of $k$,
not just asymptotically. We use the ``one-hundred plant species leaves''
data set here, because given one hundred species in this data set,
$k=100$ should be a reasonable value.

In \reffig{fig:100plants}, we vary $k$ from 2 to 200, and plot the run time
of the PAM SWAP phase \emph{only} (the cost of computing the distance matrix and
the BUILD phase is not included). We compare the original PAM, FastPAM1,
EagerPAM, and FasterPAM variants first.

\reffig{fig:100plants-opttime-lin} shows the run time in linear space,
to visualize the drastic run time differences observed.
Because the faster methods cannot be distinguished here,
we use log-log-space in \reffig{fig:100plants-opttime-log}.
Compared to the other methods, FasterPAM runtime only increases very little with $k$,
making this method particular attractive for large $k$: for $k=200$,
the FasterPAM swap run time was 169~milliseconds,
while FastPAM1 took 2.4~seconds (both using random initialization) and PAM took 151~seconds
(using, but not including, the slower BUILD initialization).

In \reffig{fig:100plants-optspeedup-log} we plot the speedup over PAM.
The FastPAM1 improvement gives an empirical speedup factor of about $0.75\cdot k$
on this particular data set,
while the additional improvements contributed an additional speedup of about
$2\text{-}7\times$ by reducing the number of iterations.
In the most extreme case tested, a speedup
of about $1190\times$ at $k=200$ for the swap procedure
(using BUILD initialization, $898\times$ with faster but worse random initialization instead)
is measured -- but because the speedup is expected to
depend on $O(k)$, the exact values are meaningless, furthermore, we excluded the distance matrix
computation and initialization in this experiment.

In \reffig{fig:100plants-opttime-norm-log}, we experimentally test the scalability in $k$
on this data set. We normalize the runtime by the parameter $k$, such that a runtime that
is linear in $k$ should approximately yield a horizontal line. While the runtime of PAM SWAP
per iteration is $O(k(n-k)^2)$, and hence one would assume a sublinear complexity, the
number of iterations increases with $k$, because PAM SWAP only performs a single swap per
iteration. EagerPAM still exhibits a runtime that increases faster than linear in $k$
(because the runtime of SWAP is still linear in $k$, and the number of iterations increases
slowly with~$k$) 
while FastPAM1 and FasterPAM empirically have sub-linear runtime in $k$ because they eliminate the
factor of $k$ within the swap iterations. We will study the dependency of the number of iterations
to the parameter $k$ in \refsec{sec:numiter}.

\begin{figure*}[t]
\begin{subfigure}{.9\linewidth}\centering
\includegraphics[width=\linewidth]{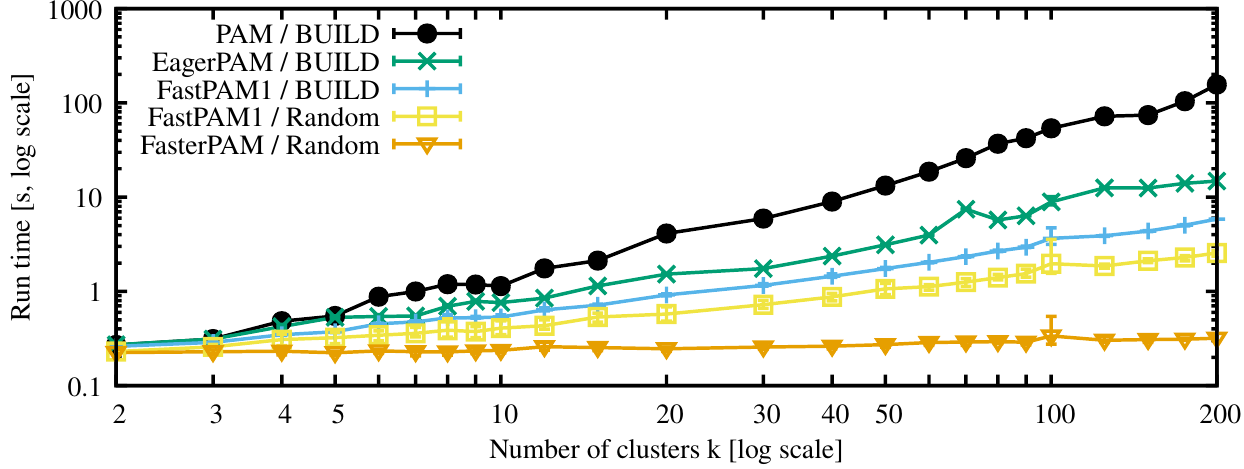}%
\vspace{-.5ex}%
\caption{Run time in log-log space}
\label{fig:100plants-runtime-log}
\end{subfigure}
\\
\begin{subfigure}{.9\linewidth}\centering
\includegraphics[width=\linewidth]{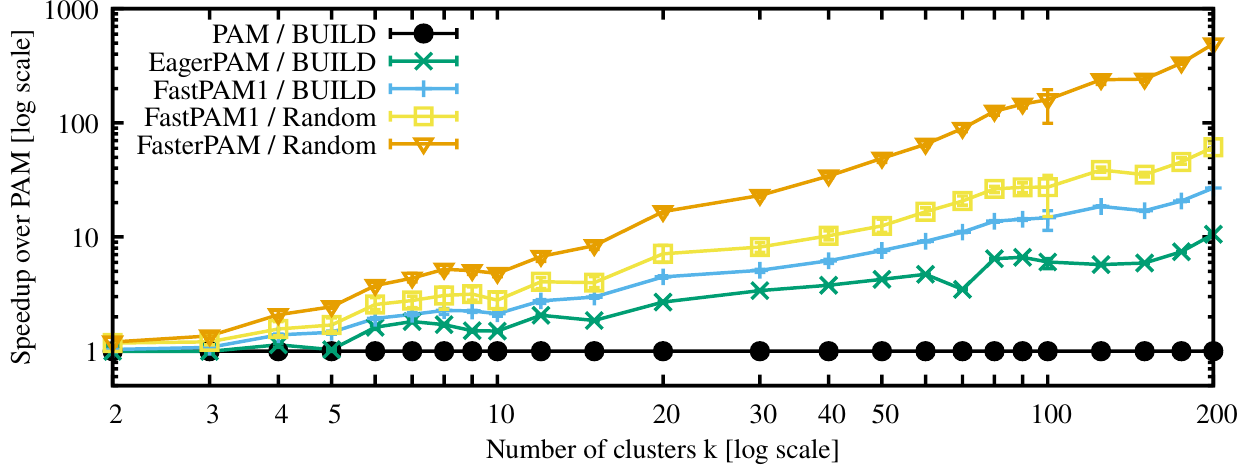}%
\vspace{-.5ex}%
\caption{Speedup in log-log space}
\label{fig:100plants-runspeedup-log}
\end{subfigure}
\caption{Run time comparison of different variations and derived algorithms.}
\label{fig:100plants-time}
\end{figure*}

If we consider the complete run time and not just the optimization,
the result does not change fundamentally: in \reffig{fig:100plants-time}
we include the distance matrix computation and initialization time.
We only present the log-log space plots, because of the extreme differences.
Because for FasterPAM a substantial amount of time is spent computing the distance
matrix, the overall run time appears to be ``almost constant''.
In the speedup factor, FastPAM1 and FasterPAM can now benefit
from using random initialization rather than BUILD
(for readability of the figures, we do not include combinations such as
FasterPAM with BUILD where runtime is dominated by BUILD).

\begin{table}[t!]\setlength{\tabcolsep}{1pt}\small
\begin{tabular}{ll|rrrrrr}
\multicolumn{2}{l|}{Algorithm  } & BUILD       & GreedyG     & LAB         &    dist.-w. & Random      & Park\&Jun \\
\hline
Initialization & time            &     2732 ms &     3251 ms &       36 ms &       14 ms &\bf     1 ms &       81 ms \\
               & relative time   &     100.0\% &     119.0\% &       1.3\% &       0.5\% &\bf    0.0\% &       3.0\% \\
               & norm. loss      &       7.4\% &\bf    3.4\% &      53.1\% &      84.6\% &      99.6\% &     274.4\% \\
               & min. n. loss    &       7.4\% &\bf    3.4\% &      48.6\% &      81.1\% &      87.3\% &     274.4\% \\
\hline
PAM            & time            &    53813 ms &\bf 37892 ms &   119640 ms &   131611 ms &   133005 ms &   146610 ms \\
               & relative time   &     100.0\% &\bf   70.4\% &     222.3\% &     244.6\% &     247.2\% &     272.4\% \\
               & norm. loss      &       0.5\% &\bf    0.1\% &       0.7\% &       0.5\% &       0.5\% &       0.4\% \\
               & min. n. loss    &       0.5\% &       0.1\% &       0.2\% &       0.2\% &\bf    0.1\% &       0.4\% \\
\hline
EagerPAM       & time            &     8910 ms &     7130 ms &     7739 ms &     6942 ms &     7048 ms &\bf  6733 ms \\
               & relative time   &      16.6\% &      13.2\% &      14.4\% &      12.9\% &      13.1\% &\bf   12.5\% \\
               & norm. loss      &       0.7\% &\bf    0.2\% &       0.6\% &       0.6\% &       0.7\% &       0.7\% \\
               & min. n. loss    &       0.7\% &       0.2\% &\bf    0.2\% &       0.2\% &       0.4\% &       0.7\% \\
\hline
FastPAM1       & time            &     3645 ms &     4232 ms &     2033 ms &\bf  1728 ms &     1967 ms &     2403 ms \\
               & relative time   &       6.8\% &       7.9\% &       3.8\% &\bf    3.2\% &       3.7\% &       4.5\% \\
               & norm. loss      &       0.5\% &\bf    0.1\% &       0.7\% &       0.5\% &       0.5\% &       0.4\% \\
               & min. n. loss    &       0.5\% &       0.1\% &       0.2\% &       0.2\% &\bf    0.1\% &       0.4\% \\
\hline
FasterPAM      & time            &     2992 ms &     3845 ms &      403 ms &\bf   321 ms &      337 ms &      444 ms \\
               & relative time   &       5.6\% &       7.1\% &       0.7\% &\bf    0.6\% &       0.6\% &       0.8\% \\
               & norm. loss      &       0.5\% &\bf    0.3\% &       0.6\% &       0.6\% &       0.6\% &       0.6\% \\
               & min. n. loss    &       0.5\% &       0.3\% &\bf    0.1\% &       0.4\% &       0.3\% &       0.6\% \\
\hline
Alternating    & time            &     3207 ms &     3352 ms &      279 ms &      252 ms &\bf   244 ms &      355 ms \\
               & relative time   &       6.0\% &       6.2\% &       0.5\% &       0.5\% &\bf    0.5\% &       0.7\% \\
               & norm. loss      &       6.9\% &\bf    3.3\% &      26.7\% &      32.9\% &      43.8\% &      96.5\% \\
               & min. n. loss    &       6.9\% &\bf    3.3\% &      20.3\% &      27.1\% &      35.0\% &      96.5\% \\
\end{tabular}
\caption{Runtime and relative loss on 100plants with $k=100$ (number of classes in this data set).
Relative time is with respect to the classic PAM algorithm with BUILD initialization,
while loss is normalized such that $0\%$ corresponds to the best solution found with
100 restarts and 100\% to the average loss of random initialization.
The minimum is the best solution of 10 restarts.}
\label{tab:100plants}
\end{table}

In \reftab{tab:100plants} we study different algorithms in combination
with different initialization methods. Similar to the experiments on the ORLib
problems, GreedyG initialization by itself already yields good results,
but is fairly expensive. In combination with PAM, GreedyG outperforms BUILD initialization.
With each algorithm, the deterministic initialization with GreedyG yields the
best result on average compared to random initialization.
Also similar to the ORLib results, the Alternating algorithm does not work very well,
and does not improve much over good starting conditions.
The approach of \citet{DBLP:journals/eswa/ParkJ09} works worst by itself, but the swapping
algorithms can still reach good results (usually at a slow runtime, though).
The proposed combination with the Alternating algorithm yields results barely better
than random initialization. EagerPAM, FastPAM1, and FasterPAM yield substantial
speedups over the original PAM. FasterPAM is fastest, as it combines both the benefits
of EagerPAM and FastPAM1. FastPAM1 yields the exact same results as PAM, but the eager
versions yield slightly worse results in this scenario.
However, this quality difference is not significant; on the ORLib data sets, the eager
variants were slightly better, and we will evaluate this in more detail in
\refsec{sec:quality}. It is worth noting, however, that the results with random
initialization can be improved by repeating the procedure multiple times. The overall
best solution in this experiment is found with PAM and FastPAM1 using random initialization:
the normalized loss is about 0.0639\%, while GreedyG initialization only found a solution of 0.0973\%.
An even slightly better solution was found with the 1000 restarts we used for normalization.
Running a fast approach (such as FasterPAM with random initialization, at 337~ms per restart;
plus the time to compute the distance matrix once of 115~ms)
multiple times will usually find a better solution faster than the deterministic solution with the best
average quality at 4232~ms (FastPAM1 with GreedyG initialization) on this data set.
While distance weighted initialization has a slight runtime advantage in this experiment,
this difference is not substantial; one may as well use uniform random initialization instead.
While LAB initialization produced better starting conditions, this would neither result
in a measurable runtime or quality advantage. Because of this, we do not present
results on LAB initialization in much detail -- it does not exhibit
a substantial advantage over the trivial random initialization.

\subsection{Number of Iterations}\label{sec:numiter}

We are not aware of theoretical results on the number of iterations needed for PAM.
The worst case may be superpolynomial like $k$-means~\citep{DBLP:conf/compgeom/ArthurV06},
albeit in practice a ``few'' iterations are usually enough.
Because of this, we are also interested in studying the number of iterations
depending on the choice of $k$ and the initialization method.

\begin{figure*}[t!]
\begin{subfigure}{.9\linewidth}\centering
\includegraphics[width=\linewidth]{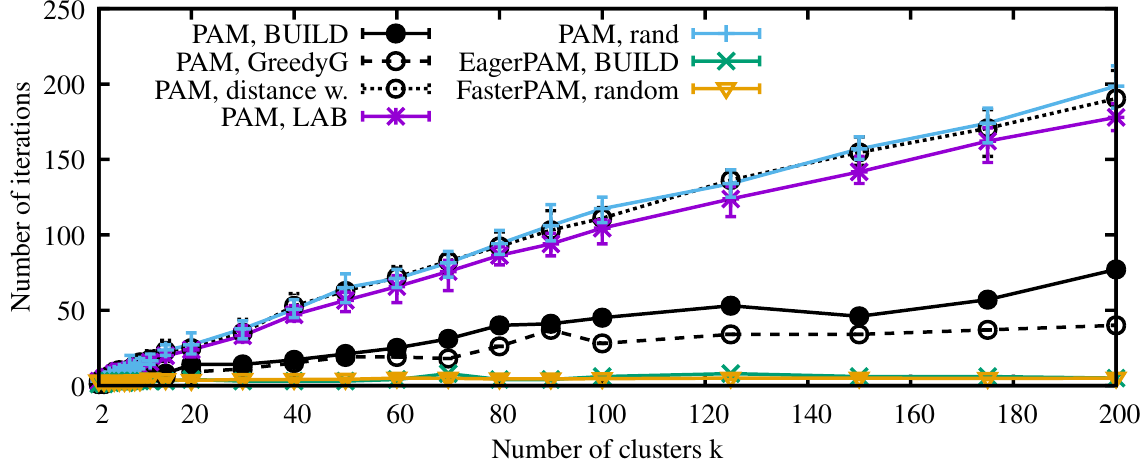}
\caption{Number of iterations until convergence}
\label{fig:100plants-iter}
\end{subfigure}
\\
\begin{subfigure}{.9\linewidth}\centering
\includegraphics[width=\linewidth]{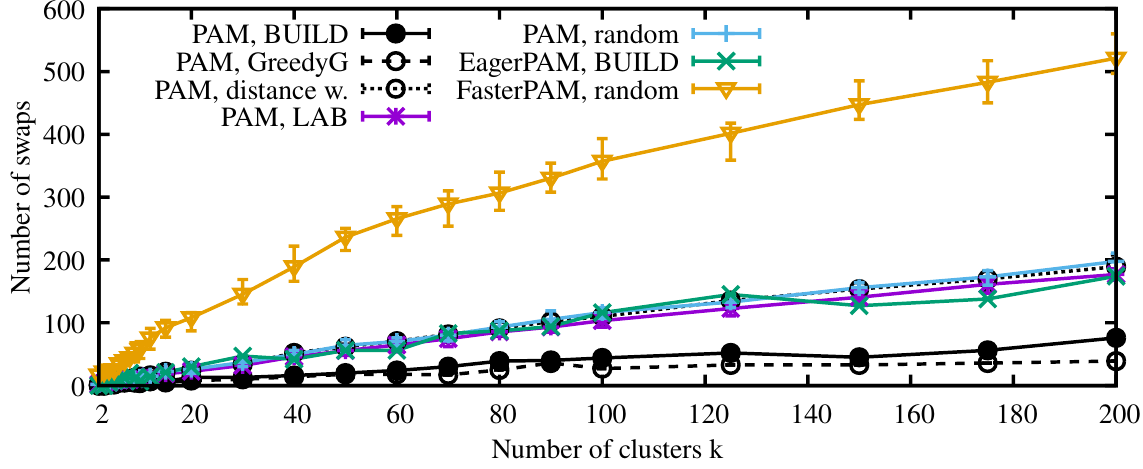}
\caption{Number of swaps performed until convergence}
\label{fig:100plants-swaps}
\end{subfigure}
\caption{Number of iterations and swaps}
\label{fig:100plants-iterswap}
\end{figure*}

For the smaller ORLib data sets, we already presented results in
\reftab{tab:init} in \refsec{sec:init}.
\reffig{fig:100plants-iterswap} shows the number of iterations needed
and the number of swaps performed with different methods.
In line with previous empirical results (e.g., \citealt{journals/infor/Whitaker83}), only ``few'' iterations are necessary.
Because PAM only performs the best swap in each iteration, a linear dependency on $k$
is to be assumed; interestingly enough we usually observed fewer than $k$ iterations
with BUILD initialization, so many medoids remain unchanged from their initial values
(note that this may be due to the small data set size). GreedyG initialization manages to reduce
the number of iterations and swaps almost by half for large $k$ compared to BUILD.
With random initialization, distance weighted and also LAB,
the number of iterations becomes approximately $k$, indicating that for almost every cluster,
a better medoid has to be chosen.
Because the distance weighted initialization requires roughly 2-4$\times$ as many iterations for PAM;
with the original algorithm where each iteration would cost
about as much as the BUILD initialization, this choice
(although suggested by \citealp{DBLP:journals/datamine/LijffijtPP15})
is detrimental even for small $k$ (c.f.{} \reftab{tab:100plants}, where PAM with distance weighted
initialization took approximately $2.45\times$ the run time of PAM with BUILD).
With the improvements of this paper, these additional iterations are cheaper than the
rather slow BUILD initialization by a factor of $O(k)$ now, hence we can now begin with
a worse but cheaper starting point.
Furthermore, because EagerPAM and FasterPAM perform multiple swaps per pass over the
data set, these two drastically reduce the number of iterations. On this data set, the
maximum number of iterations observed with EagerPAM was 13 (the worst average was 9.9),
and with FasterPAM it was 7 for a single run resp.{} 6 on average
(because for large $k$, FasterPAM performs the best of up to $k$ swaps).

We do not include the Alternating approach in these figures, because while it usually
uses the fewest iterations, it also produces substantially worse results,
as seen in \reftab{tab:100plants} and on the ORLib data. Instead we will compare
it to subsample-based algorithms such as CLARA next.

\subsection{Quality}\label{sec:quality}

\begin{figure*}[t!]
\begin{subfigure}{\linewidth}\centering
\includegraphics[width=.9\linewidth]{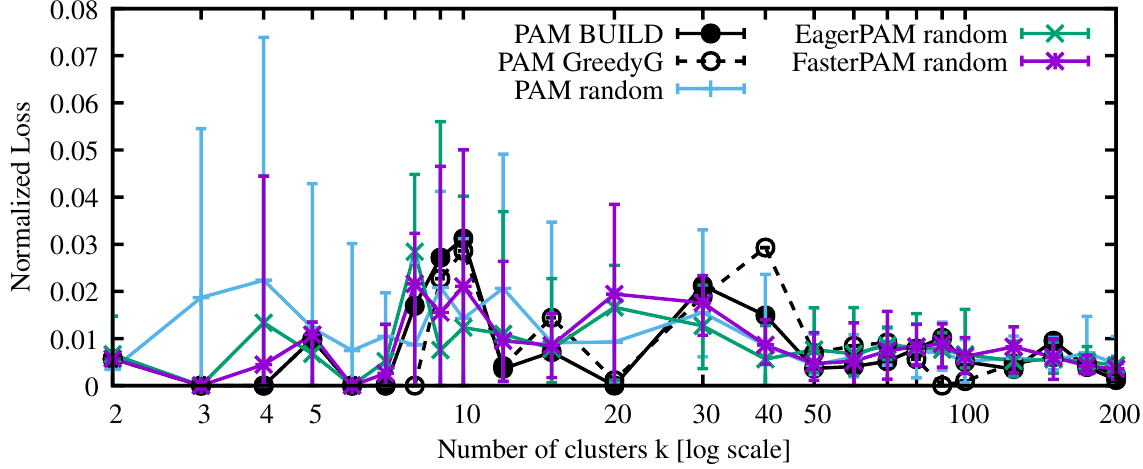}%
\vspace*{-1ex}%
\caption{Loss (\TD) with PAM variations}
\label{fig:100plants-loss-pam}
\end{subfigure}
\begin{subfigure}{\linewidth}\centering
\includegraphics[width=.9\linewidth]{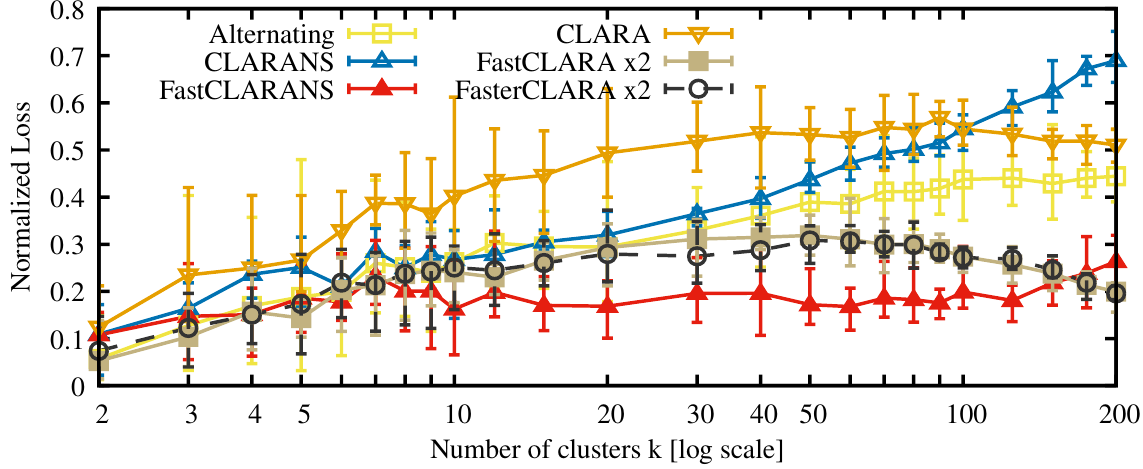}%
\vspace*{-1ex}%
\caption{Loss (\TD) of final clustering}
\label{fig:100plants-loss-approx}
\end{subfigure}
\begin{subfigure}{\linewidth}\centering
\includegraphics[width=.9\linewidth]{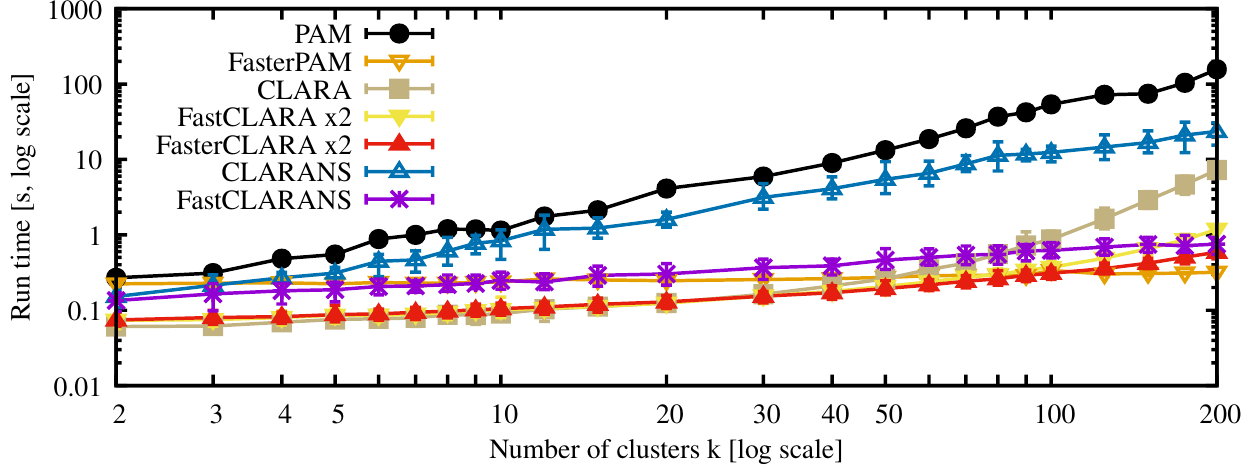}%
\vspace*{-1ex}%
\caption{Run time of approximative methods}
\label{fig:100plants-loss-runtime}
\end{subfigure}
\caption{Loss (\TD) and runtime with approximative methods}
\label{fig:100plants-loss}
\end{figure*}

Any algorithmic change and optimization comes at the risk of breaking some things,
or negatively affecting numerics (see, e.g., \citealt*{DBLP:conf/ssdbm/SchubertG18} on
how common numerical issues are, even with basic statistics such as variance in SQL databases).
In order to check for such issues, we made sure that our implementations pass the same
unit tests as the other algorithms in both ELKI and R. We do not expect numerical
problems, and PAM and FastPAM1 are supposed to give the exact same result
(and do so in the experiments, so we exclude FastPAM1 from the following plots).
EagerPAM and FasterPAM perform the first swap they find to improve the results,
and may therefore converge to a different solution. Beginning with uniform random
initialization may also yield different results. We expect that all of these
are of the same quality (because all are local optima, and neither performs
a global optimization), which we will verify experimentally.
We use the normalized loss (\refeqn{eqn:norm}), and to improve readability of the plots
we study the PAM variants separately from approximate methods such as CLARA and CLARANS.

In \reffig{fig:100plants-loss-pam} we can see that PAM and its variants (including
PAM with random initialization) find results of similar quality. Deterministic
initialization with BUILD or GreedyG is usually better than the worst solution found with random
initialization. For some $k$ such as 4 and 20, BUILD worked very well, and for $k$ such as 8 and 90 GreedyG
managed to find very good starting conditions.
But for other values of $k$ such as 10, 30, and 40 the random and eager variants were better;
GreedyG was the worst choice for $k=40$, demonstrating that it is good to try different starting conditions.
Judging from this single experiment, it may be worth considering BUILD and GreedyG for small $k<10$.
The worst of these results is still substantially better than the average performance
of any of the subsample-based methods, as seen in \reffig{fig:100plants-loss-approx} (note
the different scale on the $y$ axis).
While the Alternating algorithm considers the entire data set, it barely manages to
outperform sampling based CLARA and CLARANS in quality.
FastCLARA and FasterCLARA yield better results than CLARA because we doubled the sampling
rate; otherwise they would be similar (we omit the line from the plot for readability).
The decreasing loss of CLARA variants to the right of the plot is because the sampling size,
$40+2k$ resp.{} $80+4k$, approaches 25\% respectively 50\% of the data set size,
at which point these methods become as expensive as using the entire data set.
FastCLARANS performs best in this experiment:
compared to CLARANS it evaluates $k$ times as many possible swaps at a similar run time
cost, but as we will see later it is usually better to rather use FasterPAM instead.
As we can see in \reffig{fig:100plants-loss-runtime}, at around $k=10$, FasterPAM becomes
faster than FastCLARANS, and at around $k=100$ it outperforms also the CLARA variants.
The main benefit of CLARA is the reduced memory requirement for very large $n$,
because they do not use a full distance matrix, but instead only use one for each sample.
On a small data set such as this, they are not competitive.

\subsection{Optical Digits Dataset}\label{sec:optdigits}

\begin{table}[t!]\setlength{\tabcolsep}{1pt}
\begin{tabular}{ll|rrrrrr}
\multicolumn{2}{l|}{Algorithm  } & BUILD       & GreedyG     & LAB         & dist.-w.    & Random      & Park\&Jun \\
\hline
Initialization & time            &     6726 ms &     8326 ms &       24 ms &       28 ms &\bf     1 ms &     1202 ms \\
               & relative time   &     100.0\% &     123.8\% &       0.4\% &       0.4\% &\bf    0.0\% &      17.9\% \\
               & norm. loss      &       6.6\% &\bf    2.5\% &      39.2\% &      95.7\% &      93.5\% &     111.0\% \\
\hline
PAM            & time            &    29861 ms &\bf 23212 ms &    34837 ms &    38513 ms &    38873 ms &    35053 ms \\
               & relative time   &     100.0\% &\bf   77.7\% &     116.7\% &     129.0\% &     130.2\% &     117.4\% \\
               & norm. loss      &\bf    0.0\% &\bf    0.0\% &       0.7\% &\bf    0.0\% &       0.7\% &       1.7\% \\
\hline
EagerPAM       & time            &    17167 ms &    15936 ms &     6583 ms &\bf  6479 ms &     6534 ms &     8853 ms \\
               & relative time   &      57.5\% &      53.4\% &      22.0\% &\bf   21.7\% &      21.9\% &      29.6\% \\
               & norm. loss      &\bf    0.0\% &\bf    0.0\% &\bf    0.0\% &\bf    0.0\% &\bf    0.0\% &\bf    0.0\% \\
\hline
FastPAM1       & time            &    11493 ms &    11962 ms &\bf  6794 ms &     7329 ms &     7604 ms &     7963 ms \\
               & relative time   &      38.5\% &      40.1\% &\bf   22.8\% &      24.5\% &      25.5\% &      26.7\% \\
               & norm. loss      &\bf    0.0\% &\bf    0.0\% &       0.7\% &\bf    0.0\% &       0.7\% &       1.7\% \\
\hline
FasterPAM      & time            &     9384 ms &    10827 ms &     2277 ms &\bf  2267 ms &     2289 ms &     3618 ms \\
               & relative time   &      31.4\% &      36.3\% &       7.6\% &\bf    7.6\% &       7.7\% &      12.1\% \\
               & norm. loss      &\bf    0.0\% &\bf    0.0\% &\bf    0.0\% &\bf    0.0\% &\bf    0.0\% &\bf    0.0\% \\
\hline
Alternating    & time            &     8755 ms &     9883 ms &\bf  2023 ms &     2079 ms &     2262 ms &     3285 ms \\
               & relative time   &      29.3\% &      33.1\% &\bf    6.8\% &       7.0\% &       7.6\% &      11.0\% \\
               & norm. loss      &       3.0\% &\bf    2.5\% &       7.7\% &      20.4\% &      18.7\% &      13.6\% \\
\end{tabular}
\caption{Runtime for optdigits data with $k=10$}
\label{tab:optdigits}
\end{table}
\begin{figure}[t!]
\includegraphics[width=\linewidth]{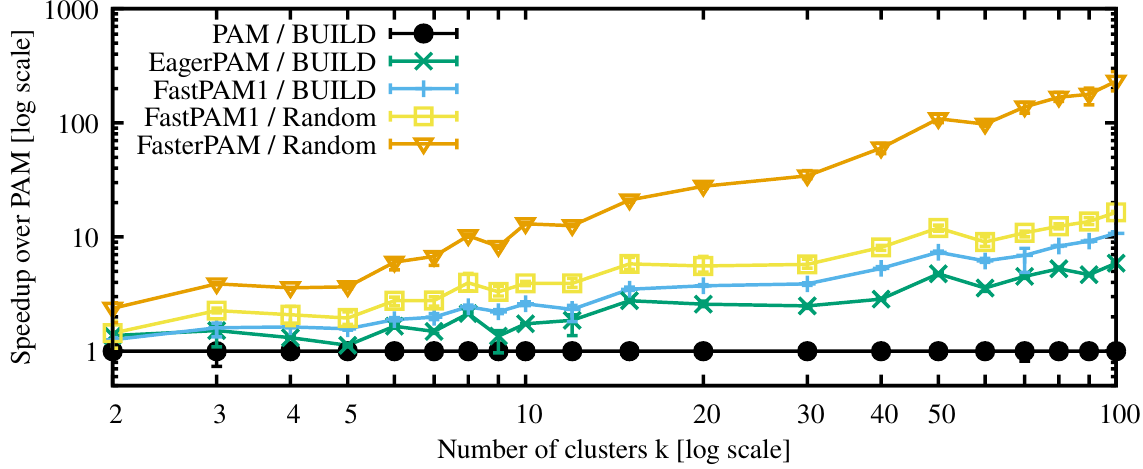}%
\vspace*{-1ex}%
\caption{Speedup on Optical Digits data}
\label{fig:optdigits-runspeedup-log}
\end{figure}

We also repeated our experiments on the slightly larger
``Optical Recognition of Handwritten Digits'' data set from UCI~\citep{Dua:2019}
with $n=5620$ instances, $d=64$ variables, and $10$ natural classes.
Example results for this data are shown in \reftab{tab:optdigits} and \reffig{fig:optdigits-runspeedup-log}.
This time, we observe better quality for the ``eager'' methods (finding the best solution
independently of the initialization), supporting the theory
that both approaches are of the same quality, and only differ in the local optima found
due to processing order. FasterPAM offers the best run time (over 10 times faster at $k=10$,
over 200 times faster total at $k=100$ as seen in \reffig{fig:optdigits-runspeedup-log}
even when including the time needed to compute the distance matrix),
and there is little benefit of using $k$-means++ or LAB over random initialization.
The Alternating approach again produces significantly worse results than the swap-based heuristics,
and does not outperform the GreedyG initialization.
Clearly, the benefits on this data set are similar.

\subsection{Scalability Experiments}\label{sec:scalability}

\begin{figure*}[t!]
\begin{subfigure}{.9\linewidth}\centering
\includegraphics[width=\linewidth]{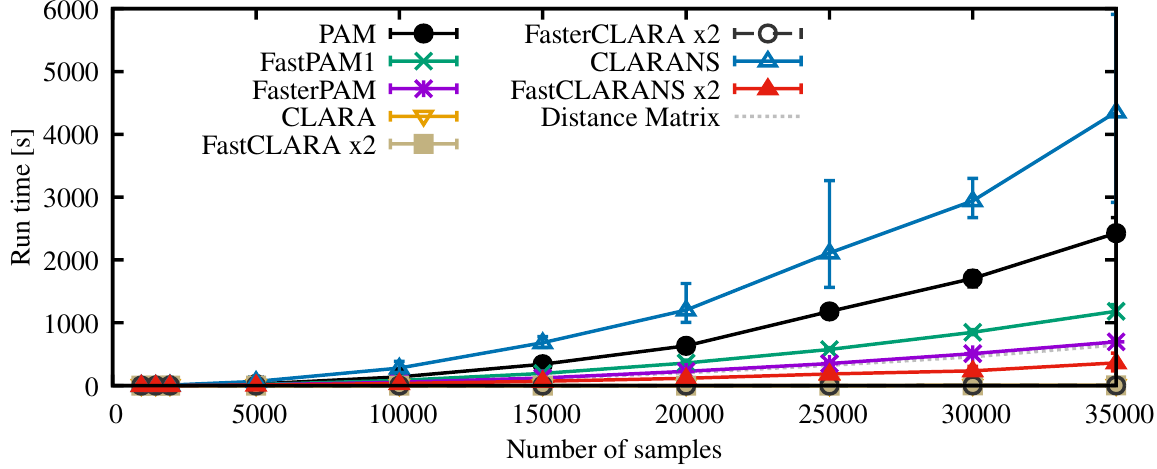}%
\vspace{-1ex}%
\caption{Run time with $k=10$}
\label{fig:mnist-runtime-10-lin}
\end{subfigure}
\hfill
\begin{subfigure}{.9\linewidth}\centering
\includegraphics[width=\linewidth]{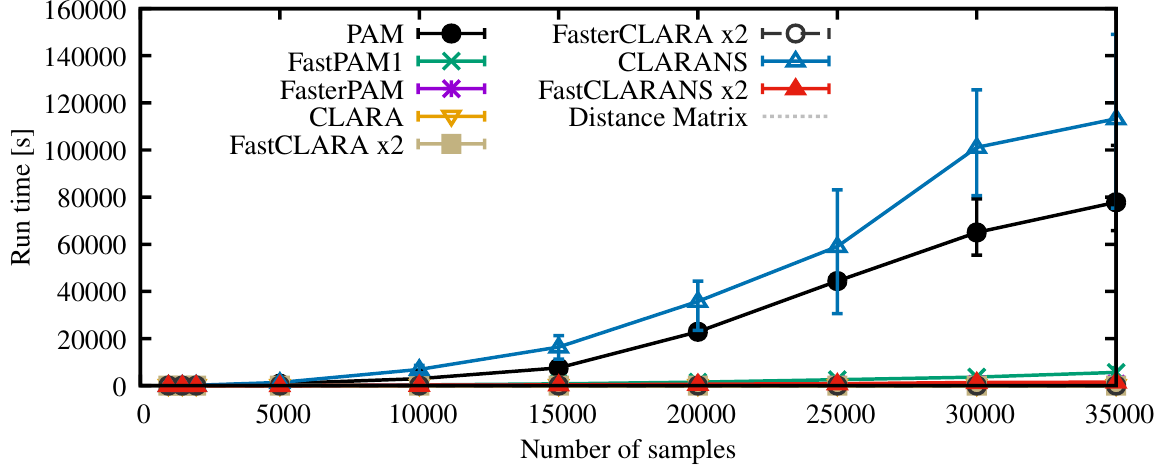}%
\vspace{-1ex}%
\caption{Run time with $k=100$}
\label{fig:mnist-runtime-100-lin}
\end{subfigure}
\caption{Run time on MNIST data}
\label{fig:mnist-runtime}
\end{figure*}

Just as PAM, our method also uses a precomputed distance matrix; the high number of distance
computations necessary makes any different use prohibitive.
This will require $O(n^2)$ time and memory, making the method as-is unsuitable for big data.
Our contributions in this paper focus on reducing the dependency on $k$, while we claim
quadratic runtime in $O(n^2)$ per iteration, and must assume the number of iterations to
potentially grow with $n$.
For larger data sets, the use of sampling-based methods such as FasterCLARA and FastCLARANS
is possible, and many scalable and distributed variations of PAM from literature can be trivially adapted to
use FasterPAM instead of~PAM.

\begin{figure*}[t!]\centering
\begin{subfigure}{.9\linewidth}\centering
\includegraphics[width=\linewidth]{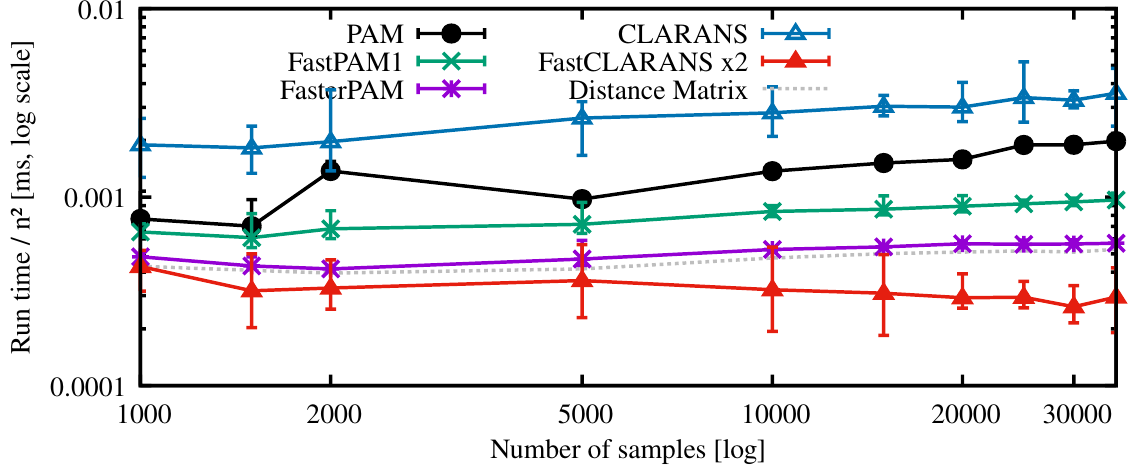}
\caption{Run time with $k=10$, normalized by $n^2$ [log-log]}
\label{fig:mnist-runtime-10-log}
\end{subfigure}
\\
\begin{subfigure}{.9\linewidth}\centering
\includegraphics[width=\linewidth]{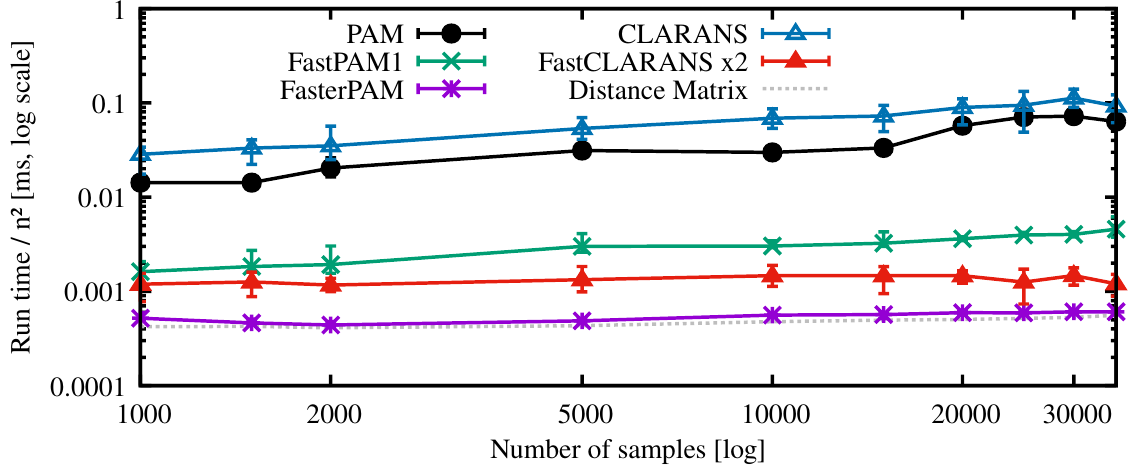}
\caption{Run time with $k=100$, normalized by $n^2$ [log-log]}
\label{fig:mnist-runtime-100-log}
\end{subfigure}
\caption{Results normalized by $n^2$ on MNIST data
with $k=10$ (top) and $k=100$ (bottom)}
\label{fig:mnist-runtime-n2}
\end{figure*}

In this experiment, we use the well-known MNIST data set from the UCI repository \citep{Dua:2019},
which has 784 variables (each corresponding to a pixel in a $28\times28$ grid) and 60000 instances.
We used the first $n=5000,10000,\ldots,35000$ instances %
and compare $k=10$ and $k=100$.
The high number of variables makes this data set expensive when that distances are recomputed
on the fly rather than using a distance matrix, as done with CLARANS, as we will see next.

The quadratic runtime growth is easily seen in the linear scale plots
\reffig{fig:mnist-runtime-10-lin}~and~\ref{fig:mnist-runtime-100-lin}.
As a reference, we give the time needed for computing the distance matrix
as dotted line, which is also quadratic.
To make the plots more interpretable, we normalize the runtime by the expected
scaling factor of $n^2$ in \reffig{fig:mnist-runtime-n2} and plot this in log-log space.
We do not include CLARA in this plot, because it subsamples the data set to a size independent of $n$.
In these plots, we can clearly see that FasterPAM is only slightly more expensive than computing
the distance matrix, and improves substantially over FastPAM1 and PAM.
CLARANS is slow in this experiment, because it does not use a distance matrix and has
to (re-)compute many distances in 784 dimensions. On low-dimensional data, it would perform better.
The authors assumed that distances are cheap to compute,
and noted that it may be necessary to cache the distances in one way or another.
For $k=10$, FastCLARANS is able to outperform the distance matrix computation (and hence all PAM variants),
and hence it clearly does not need to compute all pairwise distances. For $k=100$, it is outperformed
by the new FasterPAM approach. Hence, for expensive distance functions (on high-dimensional data,
but also, e.g., Dynamic Time Warping for time series data) and large $k$, we recommend using FasterPAM
as long as it is possible to keep a distance matrix in memory.

While the scalability in $n$ is quadratic, as expected, the most notable result is the following:
we observe that \emph{if} you can afford to compute and store the pairwise distance matrix,
then you will \emph{now} also be able to run FasterPAM. For $k=100$ and $n=35000$,
the average run time of FasterPAM was 743 seconds, of which about 655 seconds or 88\% were used
for computing the distance matrix.
The original PAM algorithm, on the other hand, took 21 hours, over 100 times longer; excluding the matrix
computation time, the speedup factor is even $879\times$. 
The main scalability problem is the memory consumption and computation of the distance matrix, not the clustering anymore.

\begin{figure*}[t!]\centering
\begin{subfigure}{.9\linewidth}\centering
\includegraphics[width=\linewidth]{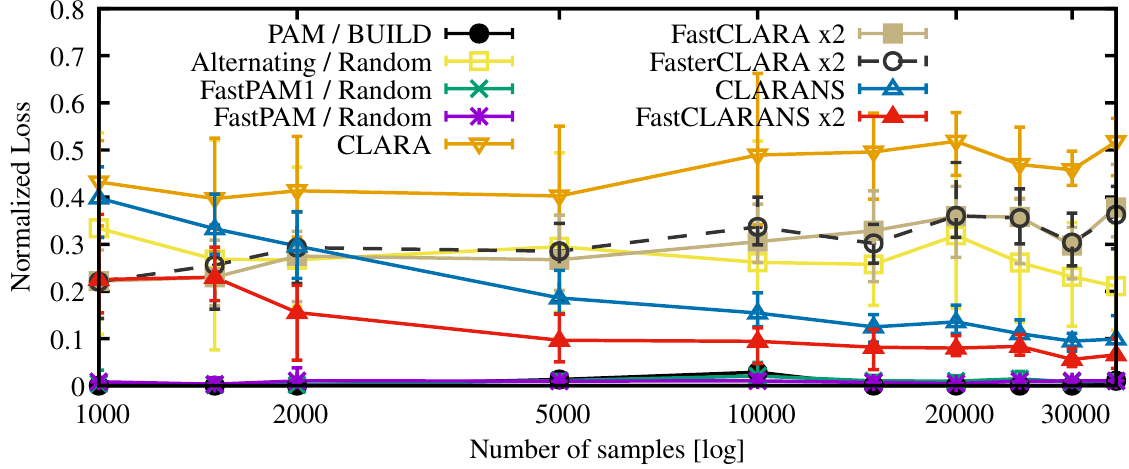}
\caption{Quality compared to the best solution found}
\label{fig:mnist-costover-10}
\end{subfigure}
\\
\begin{subfigure}{.9\linewidth}\centering
\includegraphics[width=\linewidth]{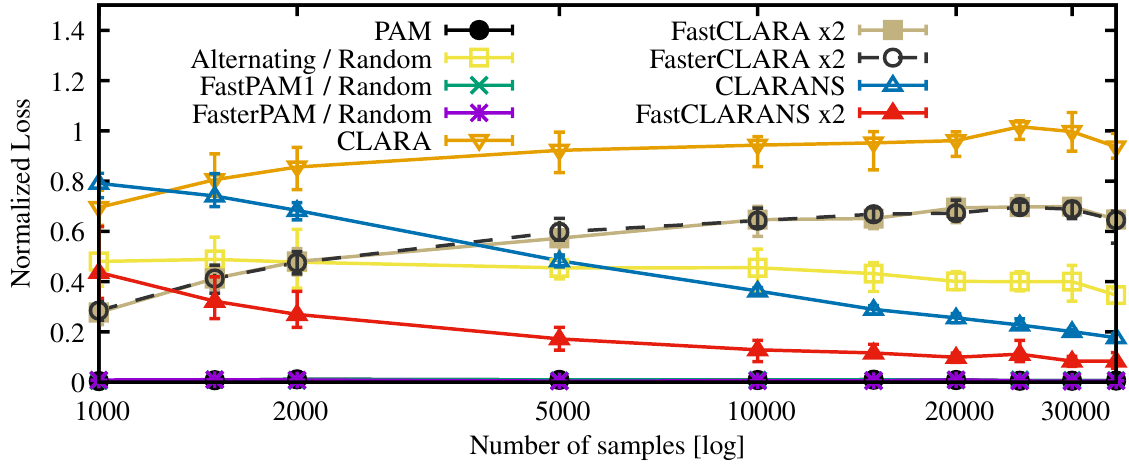}
\caption{Quality compared to the best solution found}
\label{fig:mnist-costover-100}
\end{subfigure}
\caption{Results on MNIST data %
with $k=10$ (top) and $k=100$ (bottom)}
\label{fig:mnist-quality}
\end{figure*}

If computing the distance matrix is prohibitive, it may still be possible to use
FasterCLARA (CLARA with our improved FasterPAM on the individual samples),
which will scale linearly in $n$ (in the final assignment step).
But as seen in \reffig{fig:mnist-costover-10} and
\reffig{fig:mnist-costover-100}, CLARA will usually give worse results.
Even when doubling the sampling size as suggested before, the results of
CLARA (as well as the Alternating algorithm) are about 30\% worse than
the best results found in this benchmark with $k=10$. With $k=100$
the quality of CLARA is barely better than using random medoids
(because of the low sampling rate).
The quality obtainable with CLARANS and FastCLARANS is much better.
FastCLARANS outperforms CLARANS with respect to quality because it considers
a $k$ times larger search space at a similar run time cost.
FastPAM1 and FasterPAM both yield results of the same quality as the original PAM.
But on the other hand, FastCLARANS is only advisable for inexpensive distance functions
such as (low-dimensional) Euclidean distance, and requires a non-trivial distance cache
otherwise. On complex data, we have seen that FasterPAM can be faster and give much better
results, hence FasterPAM should be the preferred method for most users.

\section{Outlook}\label{sec:outlook}

In this article we considered only the classic $k$-medoids clustering scenario,
but we have noted the close relationship to facility location problems.
Substantial research effort has been put into these problems (c.f.{} \citealt*{DBLP:journals/networks/Reese06}),
but we nevertheless hope that this research is also valuable to operations research.
The implementation used for the experiments is available as open-source,
and has already been prepared for the bichromatic case where we have separate
locations for consumers and possible suppliers, and have to find the optimal
facility placement.
We use a simple bichromatic adaptation of the medoid definition (\refeqn{eqn:def-medoid})
for possible locations~$L$ and consumers~$C$:
\begin{align*}
\operatorname{medoid}(L,C) := \argmin\nolimits_{x_l\in L} \textstyle\sum\nolimits_{x_c\in C} \dist(x_c, x_l)
\;.
\end{align*}
Other constraints popular in facility location, such as
capacity constraints and facility opening costs, are less obvious to integrate.
On the other hand, $k$-medoids clustering can likely benefit from some of this research
for clustering data without having to choose the parameter $k$ beforehand.

In the future, we also plan on working towards optimization for sparse instances,
where not all consumers can be serviced from all possible supplier locations.
Such problems arise naturally when planning or simulating for example
power networks \citep{journals/tpwrs/KaysSSWR16}, where cluster centers
correspond to power substations, consumers to households, and possible
connections are restricted to follow the road network. Because of physical
limitations such as voltage drop over distance, not every consumer can
be economically serviced from every facility location, and the procedure can be
further optimized using sparse data structures.
Given a sparse graph with $e$ edges, $n$ consumers, $m$ possible facility locations,
we assume that the algorithm presented here can be implemented in $O(e+n+m)$ time
for each iteration, which is beneficial if $e\ll n\cdot m$. However, this poses
the additional challenge of finding an admissible starting solution
(considering missing edges as infinite loss), as the problem may be unsolvable
for a small $k$. Hence, a solver for sparse problems will likely need to
also dynamically adapt $k$.

\section{Conclusions}\label{sec:conclusions}

In this article we proposed a modification of the popular PAM algorithm
that yields a provable $O(k)$ fold speedup, by clever caching of partial
results in order to avoid recomputation. 
By eagerly executing the first improvement found, we also reduced the number of
iterations substantially. We provide theoretical arguments and present
experimental evidence that this does not cause a loss in quality.
The major speedups obtained with this approach enable the use of
this classic clustering method on much larger problems, in particular with
large $k$.

Compared to our earlier work \citep{DBLP:conf/sisap/SchubertR19},
the speedup in now provable, and the eager execution approach proposed
here is both simpler and more effective than the earlier FastPAM2 method described there.
Also the initialization is much simpler now. We also given an example why
the $k$-means-style strategy yields much worse results, and should not be used.

This caching was discovered by changing the nesting order of the loops in the algorithm,
showing once more how much seemingly minor looking implementation details can
matter~\citep{DBLP:journals/kais/KriegelSZ17} and can lead to major improvements.
It is hard to devise such things on the drawing board --
such solutions more naturally arise when trying to low-level optimize the code,
such as when and when not to allocate memory for buffers, and trying to avoid
recomputing the same values repeatedly. Today's compilers are reasonably good
at performing local optimization
(at least when it does not affect numerical precision, \citealp{DBLP:conf/ssdbm/SchubertG18}),
but will not introduce an additional array to cache such values,
nor split the sums automatically.
With the faster refinement procedure of FasterPAM, it becomes possible to use cheaper initialization methods.
In contrast to our earlier work, the new experiments suggest that combining FasterPAM
with a simple uniform random initialization is often fastest and attains high quality.
We could not observe systematic improvements with either LAB or distance weighted initialization,
but the latter yield similar performance and still remain useful.

Methods based on PAM, such as CLARA, CLARANS, and the many parallel and distributed
variants of these algorithms for big data, all benefit from this improvement,
as they either use PAM as a subroutine (CLARA), or employ a similar swapping method (CLARANS)
that can be modified accordingly as seen in \refsec{sec:better-clarans}.

Our implementations are included in the open-source framework
ELKI~\citep{DBLP:journals/corr/abs-1902-03616},\footnote{\url{https://elki-project.github.io/}}
in the Rust crate ``kmedoids'',\footnote{\url{https://crates.io/crates/kmedoids}}
the Python package ``kmedoids'',\footnote{\url{https://pypi.org/project/kmedoids/}}
and the R \texttt{cluster} package\footnote{\url{https://cran.r-project.org/web/packages/cluster/index.html}}
to make it easy for others to benefit from these improvements.

With the availability in two major clustering tools, we hope
that many users will find using PAM, CLARA, and CLARANS, and later derived methods,
possible on much larger data sets with
higher $k$ than before.

\bibliographystyle{elsarticle-harv}\biboptions{authoryear}
\bibliography{fasterpam.bib}

\appendix
\section{Proof of Restructured Equation}\label{app:change-proof}

\noindent
In order to prove that \refeqn{eqn:better-change3} is equivalent to
\refeqn{eqn:td2} for all $x_o$, we decompose both equations into the
into individual contributions $\change(x_o,m_i,x_c)$ of each object $x_o$:
\begin{align*}
\Change(-m_i,x_o)=&
\begin{cases}
\dist_s(o)-\dist_n(o)
& \text{if }\nearest(o)=i
\\
0 & \text{otherwise}
\end{cases}
\\
\Change(+x_c,x_o)=&
\begin{cases}
\dist(x_o,x_c)-\dist_n(o)
&%
\text{if }\dist(x_o,x_c)<\dist_n(o)
\\
0
&%
\text{otherwise}
\end{cases}
\\
\change(x_o,m_i,x_c)=&
\Change(-m_i,x_o)+
\Change(+x_c,x_o)\notag\\&+
\begin{cases}
\dist_n(o)-\dist_s(o)
&%
\text{if }\dist(x_o,x_c)<\dist_n(o)
\text{ and }\textit{nearest}(o)=i
\\
\dist(x_o,x_c)-\dist_s(o)
&%
\text{else if }\dist(x_o,x_c)<\dist_s(o)
\text{ and }\textit{nearest}(o)=i
\\
0
&%
\text{otherwise}
\end{cases}
\end{align*}

We now prove that for all four cases of \refeqn{eqn:change}, we get the same result:

\paragraph{Case 1} If $d(x_o,x_c)<d_n(o)$, then the first and the third term cancel out
independently of $\nearest(o)$, and we always obtain $d(x_o,x_c)-d_n(o)$
from the second term.

\paragraph{Case 2} Otherwise, if $\nearest(o)=i$ and $d_n(o)\leq d(x_o,x_c)<d_s(o)$,
the second term disappears, and $d_s(o)$ cancels out, and we obtain $d(x_o,x_c)-d_n(o)$

\paragraph{Case 3} Otherwise, if $\nearest(o)=i$ and $d_s(o)\leq d(x_o,x_c)$,
the second and third terms disappear, and we get $d_s(o)-d_n(o)$ from the first.

\paragraph{Case 4} Otherwise, $d(x_o,x_c)\geq d_n(o)$ and $\nearest(o)\neq i$ must hold.
Then all terms disappear, and the result is 0.

\smallskip
Because in each case, the results are the same, 
the loss contribution for every object $x_o$ to \refeqn{eqn:better-change3}
equals \refeqn{eqn:better-change}; the sum over all $x_o$ yields the desired equality.
\hfill\qed

\end{document}